\documentclass{article}

\usepackage[preprint]{neurips_2025}

\usepackage{xstring}
\usepackage[utf8]{inputenc}
\usepackage[T1]{fontenc}
\usepackage{hyperref}
\usepackage{url}
\usepackage{booktabs}
\usepackage{amsfonts}
\usepackage{nicefrac}
\usepackage{microtype}
\usepackage{xcolor}
\usepackage{amssymb}
\usepackage{pifont}
\usepackage{colortbl}
\usepackage{xcolor}
\usepackage{amsmath}
\usepackage{amsfonts}
\usepackage{amssymb}
\usepackage{algorithm}
\usepackage{algpseudocode}
\usepackage{geometry}
\usepackage{booktabs}
\usepackage{multirow}
\usepackage{array}
\usepackage[export]{adjustbox}
\usepackage{multirow}
\newcolumntype{M}[1]{>{\centering\arraybackslash}m{#1}}

\definecolor{darkgreen}{RGB}{50,100,0}
\definecolor{darkred}{RGB}{200, 0, 0}

\usepackage{bm}
\usepackage{verbatim}
\usepackage{balance}
\usepackage{graphicx}
\usepackage{multirow}
\usepackage{xspace}
\usepackage{threeparttable}
\usepackage{enumitem}
\usepackage{makecell}
\usepackage{bbding}
\usepackage{subfig}
\usepackage{float}
\usepackage{wrapfig}
\usepackage{tabularx}
\usepackage{booktabs}
\usepackage{arydshln}
\usepackage{subcaption}
\usepackage{currfile}

\definecolor{myblue}{RGB}{0,114,189}
\newcommand{\blue}{\cellcolor{myblue!10}}

\newcommand{\cmark}{\textcolor{darkgreen}{\ding{51}}} %
\newcommand{\xmark}{\textcolor{darkred}{\ding{55}}} %
\long\def\hide#1{}

\newcommand{\model}{DeepDive\xspace}

\title{\model: Advancing Long-Horizon Search Agents with Knowledge Graphs and Multi-Turn Reinforcement Learning}

\title{\model: Advancing Deep Search Agents with Knowledge Graphs and Multi-Turn RL}

\author{%
  Rui Lu$^{12*\dagger}$, Zhenyu Hou$^{12*}$, Zihan Wang$^{23*\dagger}$, Hanchen Zhang$^{1\dagger}$, Xiao Liu$^{12}$, Yujiang Li$^{1\dagger}$, \\
  \textbf{Shi Feng$^{3}$, Jie Tang$^{1}$, Yuxiao Dong$^{1}$} \\
  \\
  $^{1}$ Tsinghua University \quad
  $^{2}$ Z.AI \quad
  $^{3}$ Northeastern University \\
  \texttt{\{learningrate1, zhenyu.hou08, wzh1998921\}@gmail.com}
}

\begin{document}

\maketitle

\renewcommand{\thefootnote}{\fnsymbol{footnote}}
    \footnotetext[1]{Equal contribution. }
    \footnotetext[2]{Work done while these authors interned at Z.AI. }
\renewcommand{\thefootnote}{\arabic{footnote}}

\begin{abstract}

Augmenting large language models (LLMs) with browsing tools substantially improves their potential as deep search agents to solve complex, real-world tasks. 
Yet, open LLMs still perform poorly in such settings due to limited long-horizon reasoning capacity with browsing tools and the lack of sufficiently difficult supervised data. 
To address these challenges, we present \model to advance deep search agents. 
First, we propose a strategy to automatically synthesize complex, difficult, and hard-to-find questions from open knowledge graphs.
Second, we apply end-to-end multi-turn reinforcement learning (RL) to enhance LLMs' long-horizon reasoning with deep search. 
To encourage diversity and reduce redundancy, we design a redundancy penalty that discourages repeated similar queries. 
Experiments show that \model-32B achieves a new open-source competitive result on BrowseComp, outperforming WebSailor, DeepSeek-R1-Browse, and Search-o1. 
We demonstrate that multi-turn RL training improves deep search ability and significantly contributes to the performance improvements across multiple benchmarks. 
We observe that \model enables test-time scaling of tool calls and parallel sampling. 
All datasets, models, and code are publicly available at \url{https://github.com/THUDM/DeepDive}. 

\end{abstract}

\begin{figure}[htbp]
  \vspace{-5pt}
  \centering
  \begin{minipage}[t]{0.31\textwidth}
    \centering
    \raisebox{4mm}{\includegraphics[width=\textwidth]{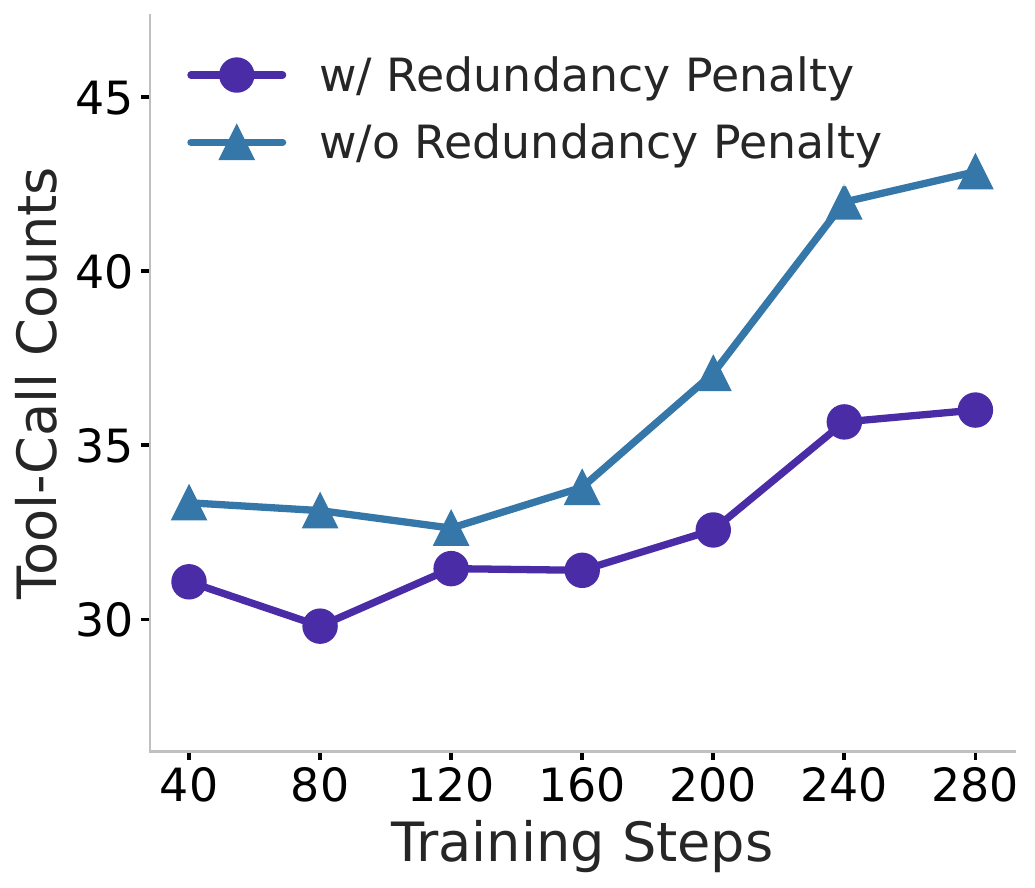}}    
  \end{minipage}
  \hfill
  \begin{minipage}[t]{0.32\textwidth}
    \centering
    \raisebox{4mm}{\includegraphics[width=\textwidth]{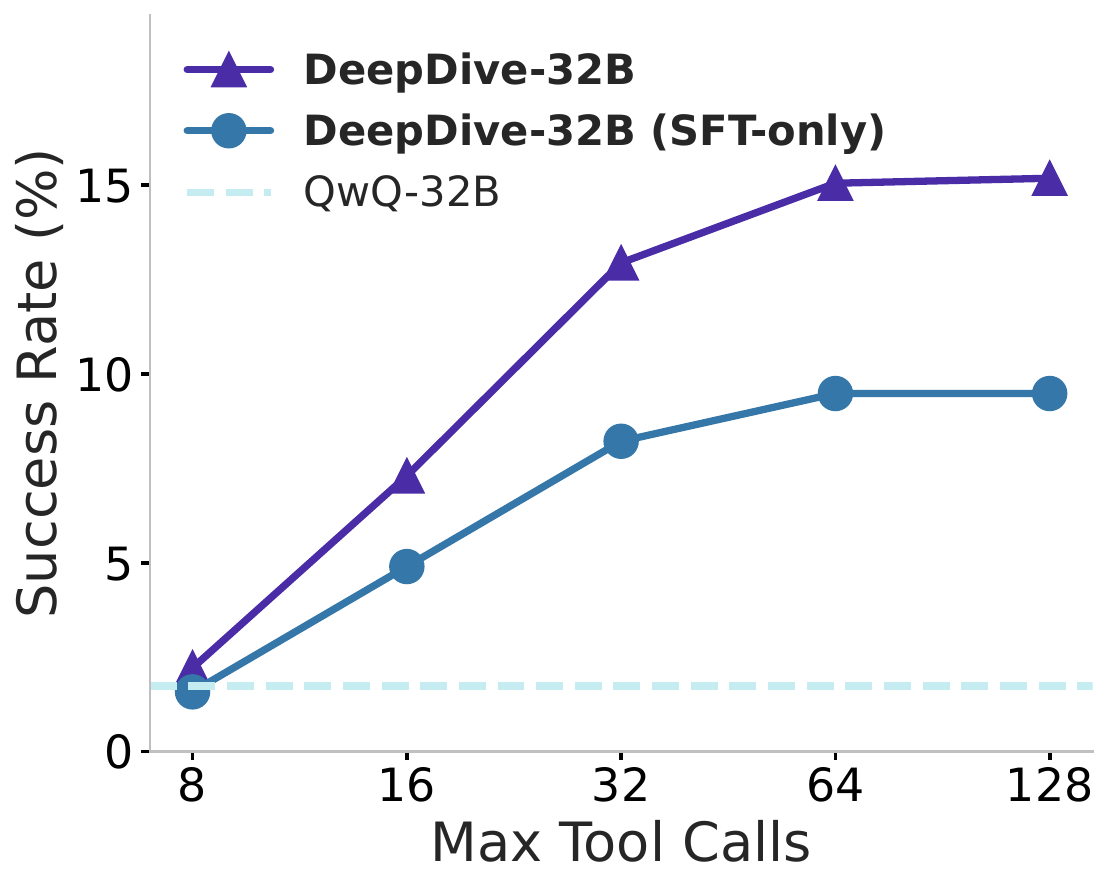}}
  \end{minipage}
  \hfill
  \begin{minipage}[t]{0.30\textwidth}
    \centering
    \includegraphics[width=\textwidth]{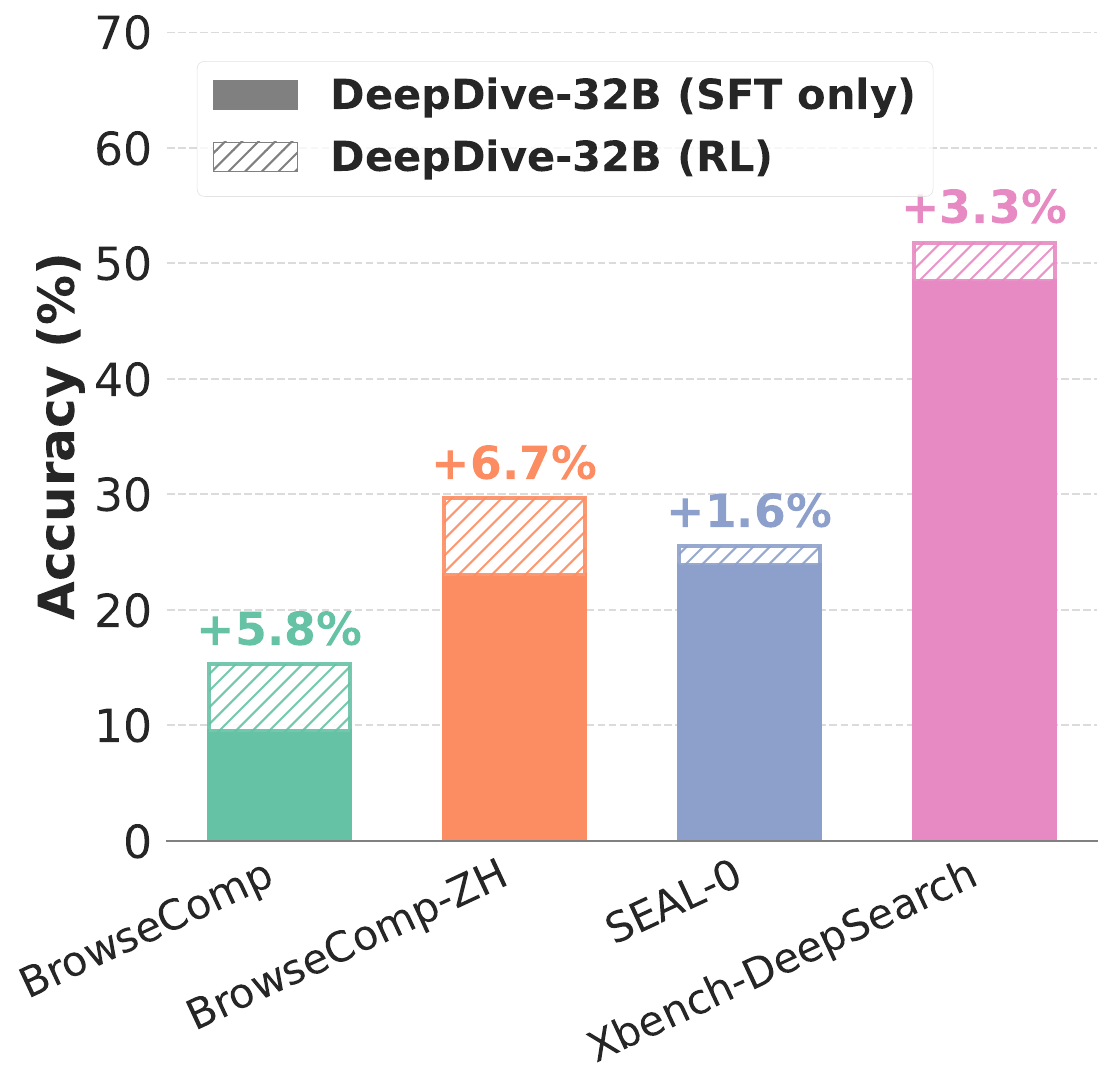}
  \end{minipage}
  \vspace{-2mm}
  \caption{%
    \textit{Left}: Adding the redundancy penalty reduces tool call counts during RL training. 
    \textit{Middle}: \model drives the model's deep search ability with maximum tool calls, which improves performance on BrowseComp. 
    \textit{Right}: Multi-turn reinforcement learning consistently enhances \model-32B on four deep search benchmarks. 
  }
  \label{fig:combined_performance}
  \vspace{-5mm}
\end{figure}

\section{Introduction}
\label{sec:intro}

Large language models (LLMs)---trained with reinforcement learning (RL) using verifiable rewards---have demonstrated strong performance in complex reasoning tasks, such as mathematics and coding competitions\citep{wei2022chain, deepseek2025r1, o1, o3, grok4}. 
As real-world tasks become increasingly complex, integrating external tools like browsing expands an LLM's knowledge beyond its training corpus. 
This shift requires the LLM to execute as an autonomous \textit{agent} capable of handing complex tasks. 

Notably, deep search agents are expected to reason over and search from hundreds of online sources to locate complex, hard-to-find information, such as answering the questions in BrowseComp \citep{wei2025browsecomp}. 
However, open models fall far behind proprietary LLMs such as OpenAI DeepResearch as deep search agents~\citep{li2025search,song2025r1,li2025webthinker,wu2025webdancer}. 
We attribute this gap to the shortage of hard-to-find data and the absence of multi-turn RL training. 

\begin{figure}[htbp]
  \centering
  \vspace{-3pt}
  \includegraphics[width=0.75\linewidth]{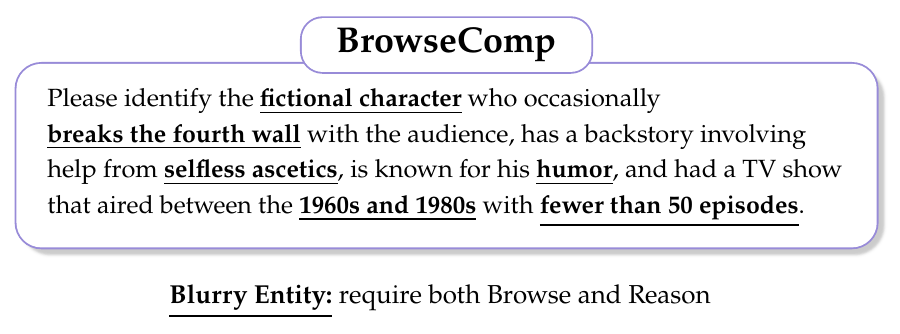}
  \vspace{-3mm}
  \caption{An illustrative example of BrowseComp \citep{wei2025browsecomp} questions, which often demand long-horizon reasoning and deep search integration across multiple blurry entities.}
  \label{fig:benchmark_comparison}
  \vspace{-4pt}
\end{figure}

First, data-wise, most existing QA datasets usually feature relatively simple questions that do not reflect true ``hard-to-find'' cases.  
For example, questions in HotpotQA~\citep{yang2018hotpotqa} can often be solved by searching for a few clear entities. 
In contrast, deep search questions such as those in BrowseComp usually involve multiple blurry entities, requiring long-horizon reasoning and deep search to reach the correct answer. 
Second, training-wise, how to effectively combine long-horizon reasoning with deep search tool use remains an open question. 
Even strong reasoning models such as DeepSeek-R1~\citep{deepseek2025r1} make only shallow tool calls and often suffer from hallucinations (see Figure \ref{fig:combined_performance} Left). 
In addition, existing browsing agents that integrate browsing tools are primarily designed to address direct search tasks. 
For example, systems like R1-Searcher \citep{song2025r1}, ReSearch \citep{chen2025research}, and DeepResearcher \citep{zheng2025deepresearcher} are mainly trained and evaluated on datasets similar to HotpotQA, including 2WikiMultiHopQA \citep{ho2020constructing}, Bamboogle \citep{press2022measuring}, and Musique \citep{trivedi2022musique}.

To address these challenges, we present \model to advance deep search agents. 
First, we automatically generate challenging QA pairs from open knowledge graphs (KGs). Second, we use end-to-end multi-turn RL to improve long-horizon reasoning in deep search scenarios. 

On the data side, we address the lack of difficulty in QA datasets by automatically constructing a deep search QA dataset from KGs, as they naturally support multi-hop connections, and each entity has different attributes. 
By deliberately blurring some attributes of each entity during question construction, we create a form of ``blurry entity''. 
We then perform random walks on the KG to extract long, multi-hop paths and use LLMs to further obfuscate key cues, making the QA pairs more challenging. 
This data synthesis process produces data that effectively stimulates LLMs' long-horizon reasoning and deep search abilities. 

On the training side, we adopt end-to-end multi-turn RL training to integrate reasoning with search tool use. 
We employ the multi-turn GRPO~\citep{shao2024deepseekmath} algorithm for RL, where the LLM interacts with a web environment and receives rewards based on the final answer in the constructed QA dataset. 
To encourage diverse exploration and prevent redundant search behavior, we design a redundancy penalty that discourages repeated similar queries as measured by Jaccard similarity. 
Figure \ref{fig:combined_performance} (middle) shows that the RL-trained model increases tool use more effectively than baselines during inference, demonstrating test-time scaling of tool calls for improved deep search.

The \model method is trained on two open models: GLM-Z1-9B-0414 \citep{glm2024chatglm} and QwQ-32B \citep{qwq32b}. 
The constructed data consists of 3,090 high-quality deep search QAs derived from KGs. 
Built on this, \model-32B reaches an accuracy of 15.3\% on  BrowseComp, surpassing many open agents---WebSailor, Search-o1, and  DeepSeek-R1-Browse---and achieving a new open-source competitive result (see Figure \ref{fig:combined_performance} left). 
Experiments demonstrate that the performance of the \model models benefits significantly from the proposed end-to-end multi-turn RL training (see Figure \ref{fig:combined_performance} right). 
We further validate across multiple challenging deep search QA benchmarks and analyze test-time scaling for tool calls and parallel sampling. 

The main contributions are summarized as follows:

\begin{itemize}[leftmargin=1.2em]
  \item We propose an automated method to synthesize deep search QA pairs from open KGs. 
  \item We introduce \model, an end-to-end multi-turn RL framework with a redundancy penalty that encourages diverse, efficient search. 
  \item Built on open models, \model-32B achieves 15.3\% on BrowseComp and shows strong test-time scaling in tool calls and parallel sampling. 
\end{itemize}

Beyond the data and method above, we also open-source an additional side study on semi-automated i.i.d. deep search QA synthesis (see details in Section \ref{sec:iid_webqa}). 
Using this data alone, the proposed end-to-end RL framework can further lift the accuracy of the 32B-parameter model to \textbf{22.2\%} on BrowseComp. 
Both the automated KG-based data and the semi-automated i.i.d. data are adopted in the open GLM-4.5 and GLM-4.6 models~\citep{team2025glm45}, contributing to their strong performance on BrowseComp.  
All the datasets, models, and code of \model are open-sourced at \url{https://github.com/THUDM/DeepDive}. 

\section{The \model Method}
\label{sec:method}

\begin{figure}[t]
  \centering
  \includegraphics[width=\linewidth]{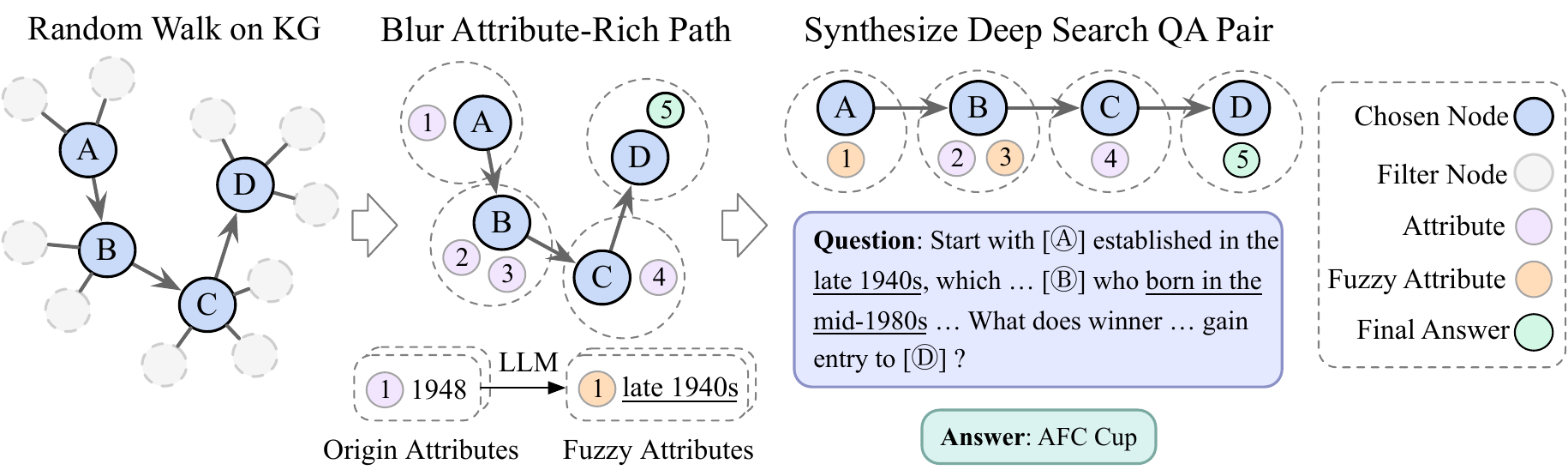}
  \caption{
    Overview of automated question–answer (QA) data synthesis from knowledge graphs (KGs) for \model. 
    Deep search QA pairs are automatically constructed by performing random walks over a knowledge graph and subsequently obfuscated using a large language model.
  }
  \label{fig:pipeline}
\end{figure}

\begin{figure}[t]
  \vspace{-5pt}
  \centering
  \includegraphics[width=0.85\linewidth]{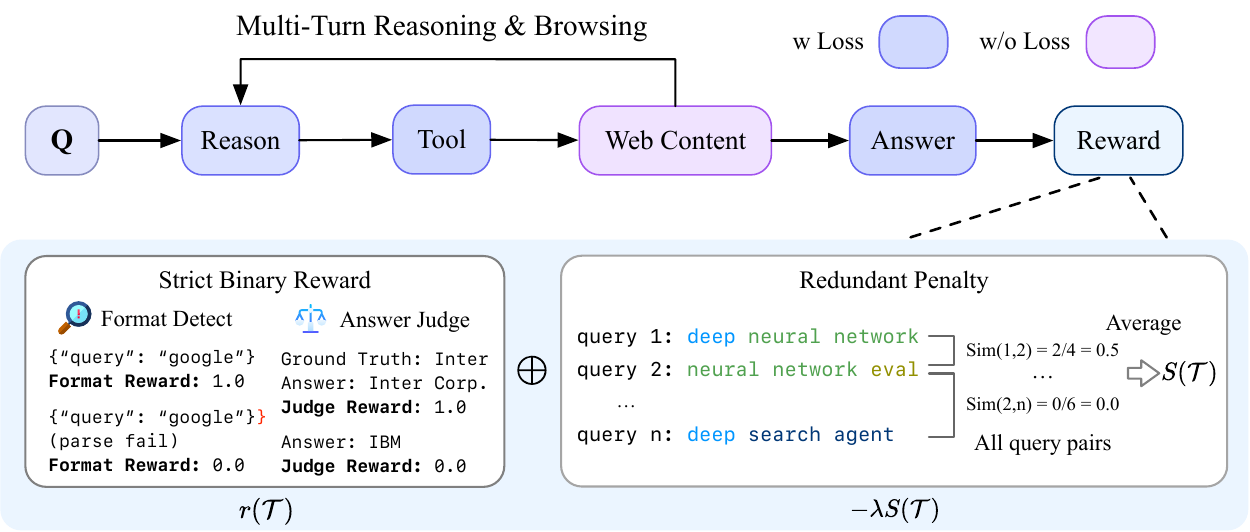}
  \caption{Overview of multi-turn RL in \model. }
  \label{fig:pipelineRL}
  \vspace{-5pt}
\end{figure}

We present \model to advance the long-horizon information-seeking ability of deep search agents. 
In \model, we introduce two techniques, targeting the data construction and RL stages, respectively. 
To generate a large-scale corpus of challenging deep-search QA pairs, we develop an automated and controllable data synthesis method with knowledge graphs (KG) (see Figure~\ref{fig:pipeline}). 
To enhance the agent's capabilities for long-horizon reasoning and browsing, we leverage the constructed data to perform end-to-end multi-turn RL training (see Figure~\ref{fig:pipelineRL}). 

To mimic human web navigation, we establish an interaction framework as the learning environment for our deep search agent. 
The agent follows an iterative cycle of reasoning, tool execution, and observation, followed by ReAct \citep{yao2023react} (see Figure~\ref{fig:pipelineRL}). 

Formally, at step $t$, the agent generates a chain-of-thought $c_t$, executes a browsing action $a_t$, and observes the web content $o_t$. 
This process repeats until the agent determines it has collected sufficient information and executes a terminating action $a_{\text{eos}}$ to give the final answer.  
The entire task execution can be represented as a trajectory $\mathcal{T}$:
\begin{equation}
    \label{eq:trajectory}
    \mathcal{T}=\left[q,\left(c_1, a_1, o_1\right), \ldots,\left(c_m, a_m, o_m\right), c_{\mathrm{ans}}, a_{\mathrm{eos}}\right], \quad m \leq n
\end{equation}
The action $a_t$ is drawn from a browsing action space with three core operations: \texttt{search}, \texttt{click}, and \texttt{open}.  
A \texttt{search} action to retrieve web page summaries with given keywords, 
a \texttt{click} action to access specific pages from search results, 
and an \texttt{open} action to access specified URLs directly.  

\subsection{Automated Data Synthesis from Knowledge Graphs}
\label{sec:data_synthesis}

Building deep search agents requires training data that goes beyond conventional multi-hop QA. 
While datasets like HotpotQA involve predictable reasoning steps, true deep search agents should act like human researchers who iteratively search, filter, and synthesize scattered evidence from the web. 
This thus calls for complex, difficult, and hard-to-find questions that even domain experts need hours to search and solve.  
Such complex training data is critical for developing agents to handle real-world tasks where information is scattered, conflicting, and hard to locate. 

However, the specific training data required to cultivate this skill is naturally scarce on the internet. 
With manual annotation being prohibitively expensive and difficult to scale, synthetic data generation emerges as the most efficient and scalable solution. 

\paragraph{Knowledge Graphs with Hard-to-Find Information.}
Naturally, knowledge graphs (KGs) provide a structured and semantically-rich environment for multi-hop reasoning, making them particularly well-suited for generating supervision data for training deep search agents. 
First, \textit{verifiability}: KGs encode factual entity-relation triples that are inherently traceable and objective, ensuring answer correctness and significantly improving data reliability compared to fully model-generated QA pairs. 
Second, \textit{multi-hop structure}: KGs allow us to explicitly control reasoning depth by performing random walks of varying lengths, enabling the generation of questions requiring multiple inference steps. 
Third, \textit{reasoning controllability}: each entity node contains multiple attributes that can be selectively obscured (such as dates, names, or locations), thereby increasing ambiguity and preventing models from exploiting shortcut solutions. 
This forces models to iteratively reason, search, and validate before finding answers. 
In light of these advantages, we propose an automated KG-based method to generate scalable, high-quality, and reasoning-intensive QA pairs.

\paragraph{Automated Data Synthesis from KGs.}
The main idea is to generate complex reasoning paths from KGs. 
A knowledge graph is a directed graph $G=(V, E)$ where $V$ represents entities and $E \subseteq V \times V$ represents relationships between them \citep{ji2021survey}. 
Each entity $v_i \in V$ has associated attributes $A(v)= \left[a_i^0, a_i^1, \cdots, a_i^t\right]$. 

To create questions that require deep reasoning and browsing, we generate paths by taking a random walk through the graph. 
Starting from an initial node $v_0$, we navigate through the graph for $k$ steps to form a path $P=\left[v_0, v_1, \ldots, v_k\right]$, where each step $(v_i, v_{i+1})$ is a valid edge in the graph. 
We choose a longer path length (e.g., $k>5$ ) to increase the potential reasoning complexity. 
However, questions generated solely based on the node sequence $P$ tend to be too simple, similar to those in HotpotQA, as their answers can be found by direct search. 

To further increase the complexity and ambiguity of the questions, we enrich and obfuscate the path by incorporating node attributes. 
Specifically, we combine each node $v_i$ in the path with its corresponding attributes to form an attribute-rich path $P_A$:
\begin{equation}
    P_A=\left[\left(v_0,\left[a_0^0, a_0^1, \ldots\right]\right),\left(v_1,\left[a_1^0, a_1^1, \ldots\right]\right), \ldots,\left(v_k,\left[a_k^0, a_k^1, \ldots\right]\right)\right]
\end{equation}
Subsequently, we select an attribute $a_k^i$ from the terminal node of the path, $v_k$, as the ground-truth answer. 
An LLM is then employed to obfuscate the information along the entire attribute-rich path $P_A$. 
This process involves techniques such as generalizing specific dates into ranges. 
The final output is a pair of challenging questions and answers $(q, a_k^i)$, generated as follows: 
\begin{equation}
    (q, a_k^i)=\text{LLM-obscure}(P_A)
\end{equation}

\paragraph{Improving Path Quality and Complexity.}
In a graph random walk, each step directly impacts the quality of the final path, which in turn determines the complexity and logical soundness of the generated QA pair. 
To improve path quality, we apply two constraints to the random walk process. 

First, we filter candidate nodes by setting an appropriate out-degree range $\left[d_{\min }, d_{\max }\right]$. 
If a node's out-degree is excessively high, it tends to be overly popular, making answers too predictable for the model. 
Conversely, nodes with low out-degree may hinder effective path expansion. 
Thus, the candidate set of nodes for the next step $\mathcal{N}\left(v_i\right)$ is defined as: 
\begin{equation}
    \mathcal{N}\left(v_i\right)=\left\{u \mid\left(v_i, u\right) \in E \wedge d_{\min } \leq d(u) \leq d_{\max }\right\}
\end{equation}

Second, to ensure logical consistency of the path, we leverage an LLM to choose the next node.  
Given the current path $P_i= \left[v_0, \ldots, v_i\right], i<k$, the LLM evaluates all candidates in $\mathcal{N}\left(v_i\right)$ and selects the most relevant next node to the existing path as $v_{i+1}$:
\begin{equation}
v_{i+1}=\text{LLM-select}\left(P_i, \mathcal{N}\left(v_i\right)\right)
\end{equation}

Together, these constraints guide the random walk to produce reasoning paths that are both complex and coherent, synthesizing high-quality QA pairs. 

To further increase question difficulty, we implement an automated filter using a frontier model (e.g., GPT-4o~\citep{gpt4o}) with basic search capabilities. 
Each question is tested four times—if the model solves it in any attempt, the question is discarded. 
Only questions that fail all four attempts are retained, ensuring our dataset contains exclusively challenging tasks requiring complex reasoning and advanced web browsing rather than simple information lookups. 

\subsection{End-to-End Multi-Turn Reinforcement Learning}

Given the challenging QA dataset, we use end-to-end multi-turn reinforcement learning (RL) to train the agent for deep search. 
Based on the standard GRPO algorithm for multi-turn RL, we enhance the reward mechanism by combining strict rewards for correctness with a redundancy penalty to encourage search diversity. 

\paragraph{Multi-Turn RL.}
Unlike single-turn RL, where the model outputs a single response per question, multi-turn RL lets the agent perform multiple reasoning and tool-use steps before arriving at a final answer. 
We employ the Group Relative Policy Optimization (GRPO) algorithm \citep{shao2024deepseekmath} to train the deep search agent. 
For each question $q$, we sample the tool calling trajectories $G$ from the current policy $\pi_{\theta}$. 
For each trajectory $\mathcal{T}$, we then calculate a normalized advantage $A_i=\left(r_i-\operatorname{mean}\{r_k\}_{k=1}^G\right) / \operatorname{std}\left\{r_k\right\}_{k=1}^G$, then the policy parameters $\theta$ are updated to maximize a clipped objective function with a KL penalty: 
\begin{equation}
    \label{eq:rl_object}
    \mathcal{L}(\theta)=\frac{1}{G} \sum_{i=1}^G\left[\min \left(\rho_i A_i, \operatorname{clip}\left(\rho_i, 1-\epsilon, 1+\epsilon\right) A_i\right)-\beta \mathrm{KL}\left(\pi_\theta \| \pi_{\mathrm{ref}}\right)\right]
\end{equation}
This objective uses the importance ratio $\rho_i=\pi_\theta(\mathcal{T}) / \pi_{\theta_{\text {old }}}(\mathcal{T})$, where $\epsilon$ controls the clipping range and $\beta$ weights the penalty for diverging from a reference policy $\pi_{\text{ref}}$. 

\paragraph{Encouraging Diverse Search with Redundancy Penalty. }
Deep search tasks are inherently multi-turn, as formalized in Eq.~\ref{eq:trajectory}. 
Deep search tasks benefit significantly from diverse exploration strategies, as different search queries can uncover complementary information and lead to a more comprehensive understanding. 
To promote such diversity, we design a reward mechanism that encourages browsing agents to explore varied search approaches while maintaining correctness. 

Our approach combines two key components. 
First, we measure search diversity by analyzing how similar queries are within a search trajectory. 
Given a trajectory $\mathcal{T}$ with all search queries $Q=[q_1, q_2, \ldots, q_T]$, where each query $q_i$ contains keywords $q_i=\{w_{i,1}, w_{i,2}, \ldots, w_{i,n_i}\}$, we calculate the Jaccard similarity \citep{real1996probabilistic} between any two queries as: $\operatorname{sim}(q_i, q_j) = |q_i \cap q_j| / |q_i \cup q_j|$. 
The overall similarity across all queries in the trajectory is then computed as: 
\begin{equation}
    S(\mathcal{T}) = \frac{1}{T(T-1)} \sum_{i \neq j} \operatorname{sim}(q_i, q_j),\quad S(\mathcal{T}) \in [0,1]
\end{equation}
This metric equals 1 when all queries are identical and 0 when all queries are completely disjoint. Lower similarity indicates more diverse search exploration. 

Second, we employ a strict binary reward to ensure trajectory correctness. 
A trajectory $\mathcal{T}$ receives a $+1$ reward only when every step is correctly formatted, including the reason $c_i$ and the action $a_i$, and the final answer $a_{\text{eos}}$ matches the ground-truth $a^*$. 
Since entities may have multiple valid representations, we use an LLM judge \citep{zheng2023judging} for answer verification. Formally, the binary reward is defined as: 
\begin{equation}
    \label{eq:reward}
    r(\mathcal{T})=
    \begin{cases}
    1,&\left(\forall\ i, \text{Format}\left(c_i, a_i\right)\right) \wedge \text{Judge}\left(a_{\mathrm{eos}}, a^*\right) \\
    0,& \text{otherwise}
    \end{cases}
\end{equation}

We combine these components into our final reward function: 
\begin{equation}
    r'(\mathcal{T}) = r(\mathcal{T}) - \lambda \cdot S(\mathcal{T})
\end{equation}
where $\lambda < 1$ controls how much we reward diverse queries. 
This formulation encourages agents to explore a wider range of search strategies while maintaining a strong emphasis on the correctness of the final answer, thereby fostering more efficient and comprehensive search behaviors. 
\section{Experiments}
\label{sec:experiment}

\subsection{Setup}
\label{sec:exp_setup}

\paragraph{Benchmarks. }
We evaluate \model on four public and challenging deep search benchmarks: BrowseComp \citep{wei2025browsecomp}, BrowseComp-ZH \citep{zhou2025browsecomp}, Xbench-DeepSearch \citep{xbench}, and SEAL-0 \citep{pham2025sealqaraisingbarreasoning}. 

\paragraph{Baselines. }
\label{sec:baselines}
We compare \model against a diverse set of models, grouped into two categories:
\textbf{(1) Proprietary models:} 
This group includes both non-browsing and browsing-capable models. Non-browsing models consist of GPT-4o \citep{gpt4o}, Claude-3.7-Sonnet \citep{claude37}, Claude-4-Sonnet-Thinking \citep{anthropic2025extended-thinking} and o1 \citep{o1}, which are evaluated solely on their internal reasoning abilities. Browsing-capable proprietary models include Grok-DeepResearch \citep{grok}, Doubao with Deep Think and Search \citep{doubao}, and OpenAI’s Deep Research \citep{openai2025deepresearch}. Additionally, we extend select non-browsing models with our browsing tools to examine performance gains via standard function calls. 
\textbf{(2) Open-source models:} 
This group includes recent high-performing open-source models, both with and without browsing capabilities. 
The non-browsing models consist of GLM-Z1-9B-0414 \citep{glm2024chatglm}, DeepSeek-R1-0528 \citep{guo2025deepseek} and QwQ-32B \citep{qwq32b}. 
We compare our method with recent open-source web agents, including Search-o1 \citep{li2025search}, ASearcher \citep{gao2025beyond}, WebDancer \citep{wu2025webdancer}, and WebSailor \citep{li2025websailornavigatingsuperhumanreasoning}. 
To ensure a fair comparison, we also enable standard function calling for GLM-Z1-9B-0414 and QwQ-32B, allowing them to browse during evaluation. 

\paragraph{Data Synthesis Details. }
We build synthetic datasets from two public knowledge graphs, KILT \citep{petroni2020kilt} and AMiner\citep{tang2012arnetminer}. 
First, we generate long-chain paths via random walking with parameters set to $k \in[5,9]$, $d=3, d_{\min}=4, d_{\max}=8$. 
We then use Gemini-2.5-Pro \citep{team2023gemini}, leveraging its superior long-context ability, to obscure entities and synthesize the QA pairs. 
This process yields 3,250 deep search QA pairs, which are randomly split into 1,016 samples for Supervised Fine-Tuning (SFT) and 2,234 for Reinforcement Learning (RL). 

\paragraph{Training Details. }
We integrate the Serper API \citep{serper2025} for web search, which returns the top-10 pages for each query. 
The Jina API \citep{jina} handles the click and open operations. 
Our training process follows recent RL approaches for large language models \citep{guo2025deepseek, hou2025advancing, li2025websailornavigatingsuperhumanreasoning}, starting with a cold-start phase. 
We leverage the Claude-4-Sonnet-Thinking model \citep{anthropic2025extended-thinking}, which has tool-calling capabilities, to interact with browsing tools and generate cold-start data through multiple attempts and reject sampling, yielding 858 high-quality SFT traces. 

We choose two open models as our backbone models: GLM-Z1-9B-0414 \citep{glm2024chatglm} and QwQ-32B \citep{qwq32b}. 
Each model is trained for 3 epochs with a global batch size of 32, a learning rate of $1\times10^{-5}$, and a maximum context length of 104,800.  

During RL, we conduct training using the open-source Slime framework \citep{slime_github} with all 2,234 data samples. 
The training configuration includes a rollout size of 8, 16 samples per prompt, a global batch size of 128, a temperature of 1.0, and a maximum context length of 51,200 tokens. 
We set the redundancy penalty coefficient to $\lambda=0.1$. 
To promote exploration, we set the KL penalty coefficient to $\beta = 0$ \citep{Vassoyan2025IgnoreKL} and employ a learning rate of $1 \times 10^{-6}$. 

\paragraph{Evaluation. }
For datasets and models with previously-reported scores, we directly adopt the results from their respective papers. 
For all other evaluations, we follow the LLM-as-Judge framework \citep{zheng2023judging}, employing Llama-3.1-70B \citep{llama3} to assess whether a model’s final output matches the ground truth answer. 
To speed up evaluation during reinforcement learning (RL) training, each checkpoint is assessed on a fixed, randomly pre-sampled subset of BrowseComp-266, with a maximum of 75 turns. 
Once training saturates, we evaluate later checkpoints on the full BrowseComp dataset (1,266 instances) with the turn limit raised to 128.
For other benchmarks whose total size is below 300, we evaluate on the entire dataset. To reduce variance and improve robustness, every dataset is evaluated twice, and the average accuracy is reported as the final result.

\subsection{Overall Performance}

\begin{table}[!ht]
  \centering
  \caption{
      Evaluation of deep search QA benchmarks. 
      Accuracy(\%) is reported. 
      * represents reported performance from existing studies. 
      \textbf{bold}: best among open-source models; \underline{underline}: second best. 
  }
  \label{tab:main-result}
  \renewcommand{\arraystretch}{1.3}
  \begin{adjustbox}{max width=\textwidth}
    \begin{tabular}{lcccccc}
      \toprule
      \textbf{Model} & \textbf{Reason} & \textbf{Browse} &
      \textbf{BrowseComp} & \textbf{BrowseComp-ZH} & \textbf{Xbench-DeepSearch} & \textbf{SEAL-0} \\
      \midrule
      \multicolumn{7}{c}{\textbf{\textit{Proprietary Models}}} \\
      \midrule
      GPT-4o              & \xmark & \xmark & 0.9* & 11.1 & 18.0* & 0.9 \\
      GPT-4o             & \xmark & \cmark & 1.9* & 12.8 & 30.0 & 9.1 \\
      Claude-3.7-Sonnet              & \xmark & \xmark & 2.3 & 11.8 & 12.0 & 2.7 \\
      Claude-3.7-Sonnet         & \xmark & \cmark & 4.5 & 14.2 & 29.0 & 14.4 \\
      o1              & \cmark & \xmark & 9.9* & 29.1* & 38.0 & 11.7 \\
      o4-mini              & \cmark & \xmark & 6.1* & 15.2* & 22.3* & 2.7 \\
      Claude-4-Sonnet-Thinking         & \cmark & \xmark & 2.6 & 21.5 & 27.0 & 9.0 \\
      Claude-4-Sonnet-Thinking        & \cmark & \cmark & 14.7 & 30.8 & 53.0 & 37.8 \\
      Grok-DeepResearch    & \cmark & \cmark & - & 12.9* & 50.0* & - \\
      Doubao-DeepThink    & \cmark & \cmark & - & 26.0* & \underline{50+} & - \\
      DeepResearch & \cmark & \cmark & 51.5* & 42.9* & - & - \\
      \midrule
      \multicolumn{7}{c}{\textbf{\textit{Open-Source Models}}} \\
      \midrule
      GLM-Z1-9B-0414              & \xmark & \xmark & 0.6 & 2.4 & 8.0 & 7.2 \\
      GLM-Z1-9B-0414              & \xmark & \cmark & 0.6 & 1.7 & 3.0 & 2.7 \\
      QwQ-32B                           & \cmark & \xmark & 1.7 & 13.5 & 10.7* & 5.4 \\
      QwQ-32B           & \cmark & \cmark & 1.3 & 14.5 & 27.0 & 4.5 \\
      DeepSeek-V3-0324                  & \xmark & \xmark & 1.5 & 24.6 & 36.0 & 6.3 \\
      DeepSeek-R1                       & \cmark & \xmark & 2.0 & 23.2 & 32.7* & 5.4 \\
      DeepSeek-R1-0528                  & \cmark & \xmark & 3.2 & 28.7 & 37.0 & 5.4 \\
      GLM-4.5-Air-106B & \cmark & \cmark & 21.3 & 36.3 & 65.0 & 30.6 \\
      GLM-4.5-355B & \cmark & \cmark & 26.4 & 37.5 & 68.0 & 36.0 \\
      \midrule
      Search-o1-32B     & \cmark & \cmark & 2.8* & 17.9* & 25.0* & - \\
      ASearcher-Web-32B & \cmark & \cmark & 5.2* & 15.6* & 42.1* & - \\
      WebThinker-32B     & \cmark & \cmark & 2.8* & 7.3* & 24.0* & - \\
      WebDancer-32B     & \cmark & \cmark & 3.8* & 18.0* & 39.0* & - \\
      WebSailor-7B     & \cmark & \cmark & 6.7* & 14.2* & 34.3* & - \\
      WebSailor-32B     & \cmark & \cmark & \underline{10.5}* & \underline{25.5}* & \textbf{53.3}* & - \\
      \midrule
      \blue{\model-9B (\textit{sft-only})} & \blue{\cmark} & \blue{\cmark} & \blue{5.6} & \blue{15.7} & \blue{35.0} & \blue{15.2} \\
      \blue{\model-9B} & \blue{\cmark} & \blue{\cmark} & \blue{6.3} & \blue{15.1} & \blue{38.0} & \blue{12.2} \\
      \blue{\model-32B (\textit{sft-only})} & \blue{\cmark} & \blue{\cmark} & \blue{9.5} & \blue{23.0} & \blue{48.5} & \blue{\underline{23.9}} \\
      \blue{\model-32B} & \blue{\cmark} & \blue{\cmark} & \blue{\textbf{15.3}} & \blue{\textbf{29.7}} & \blue{\underline{51.8}} & \blue{\textbf{25.5}} \\
      \bottomrule
    \end{tabular}
  \end{adjustbox}
\end{table}

Table~\ref{tab:main-result} presents a comprehensive comparison between \model and a range of baselines across four challenging deep search benchmarks. From the results, we draw the following key observations: 

\paragraph{Competitive among Open Deep Search Agents. }
The \model-32B model excels on four challenging deep search benchmarks. 
For the BrowseComp benchmark, it ranks just behind OpenAI’s DeepResearch and far ahead of other open-source models or agents. 
While most open-source models score under 10\% on BrowseComp, \model-32B achieved 15.3\%. 
It also shows clear advantages on SEAL-0 and XBench-DeepSearch, indicating effective use of browsing for complex reasoning. 
The results also highlight the power of reinforcement learning (RL). 
The 32B model with only SFT has already scored 9.5\% on BrowseComp, RL then enhances the model’s core ability to combine reasoning with search, resulting in stable performance growth over the SFT version on each benchmark. 
Notably, these gains are less pronounced for the smaller 9B model, potentially because of its limited reasoning capacity or a tendency to overfit on synthetic data during its training. 

\paragraph{RL Drives Deeper Search Strategies. }

\begin{wrapfigure}{r}{0.5\textwidth}
  \centering
  \vspace{-10pt}
  \includegraphics[width=0.5\columnwidth]{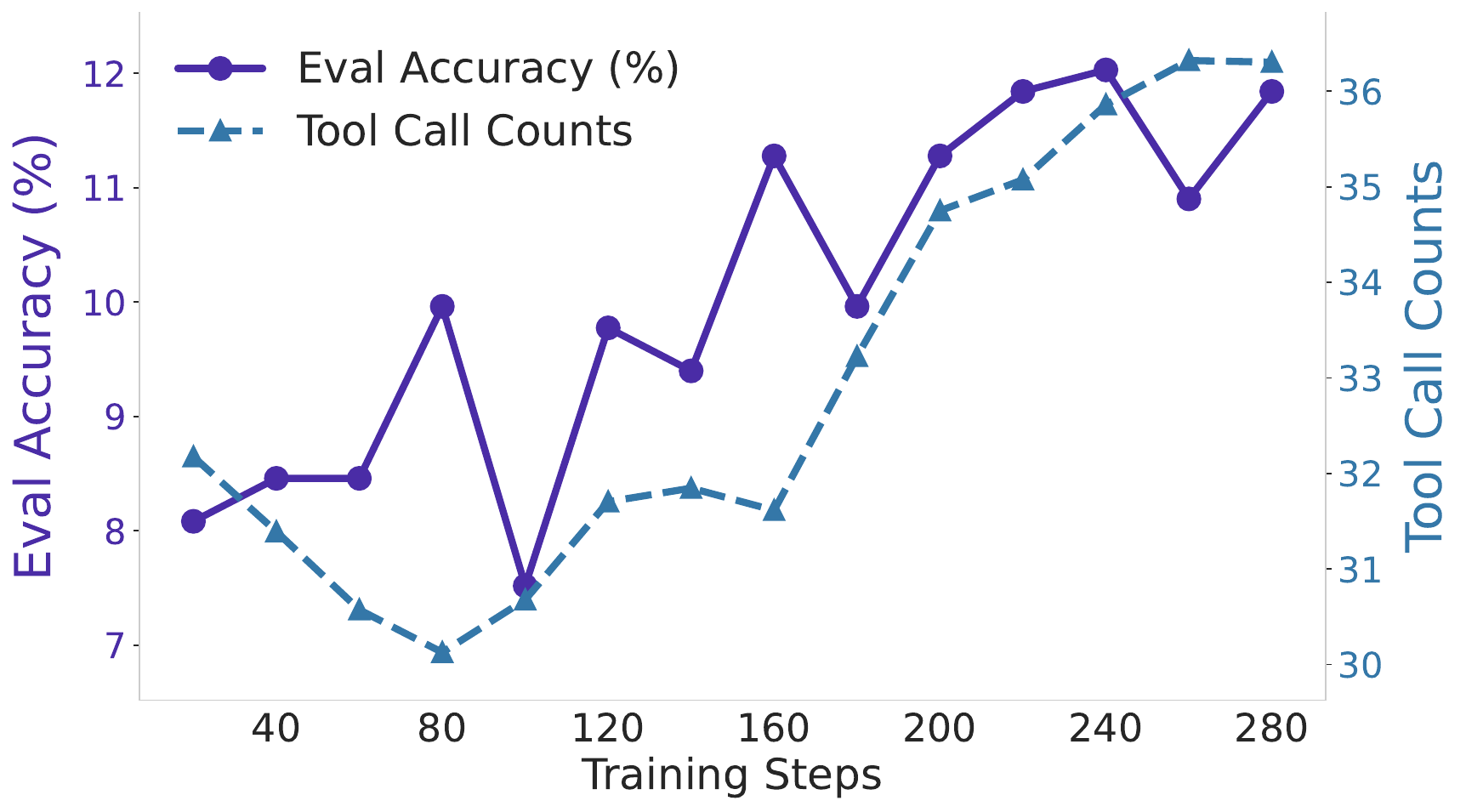}
  \caption{Evaluation accuracy and tool calls during RL training on a random subset (BrowseComp-266). }
  \label{fig:browsecomp_266_reward}
  \vspace{-20pt}
\end{wrapfigure}

Figure~\ref{fig:browsecomp_266_reward} illustrates the effect of reinforcement learning on \model-32B through two key metrics: model performance and tool call counts. 
Evaluation accuracy on a randomly sampled subset (BrowseComp-266) consistently improves, accompanied by rising tool usage, indicating that the model explores progressively deeper search strategies. 
These results demonstrate that reinforcement learning trained on our synthetic data successfully enhances both performance and search depth, with benefits generalizing to unseen samples. 

\subsection{Test-Time Scaling for \model}

We evaluate the test-time scaling capabilities of our model from two perspectives: single attempt scaling by increasing the tool call budget, and multiple attempt scaling through parallel sampling with different answer selection strategies. 
These experiments demonstrate how additional computation at inference time can substantially improve model performance. 

\paragraph{Tool Call Scaling during Inference. }

\begin{figure*}[!htbp]
    \vspace{-10pt}
    \centering
    \subfloat[BrowseComp]{
        \begin{minipage}{0.32\textwidth}
            \centering
            \includegraphics[width=\textwidth]{./figure/browsecomp-serper-scaling-log.pdf}
            \vspace{-3mm}
            \label{fig:browsecomp_scaling}
        \end{minipage}
    }\hfill
    \subfloat[BrowseComp-ZH]{
        \begin{minipage}{0.32\textwidth}
            \centering
            \includegraphics[width=\textwidth]{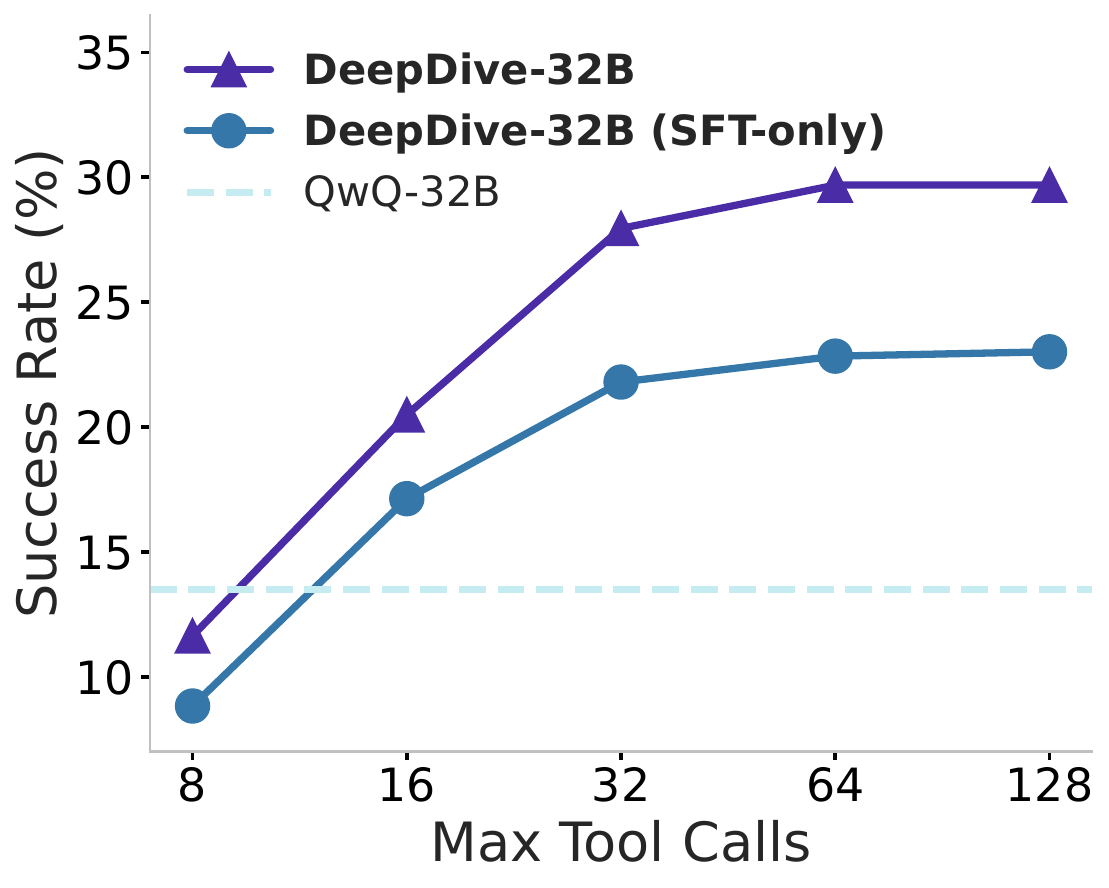}
            \vspace{-3mm}
            \label{fig:browsecomp_zh_scaling}
        \end{minipage}
    }\hfill
    \subfloat[Parallel Sampling]{
        \begin{minipage}{0.32\textwidth}
            \centering
            \includegraphics[width=\textwidth]{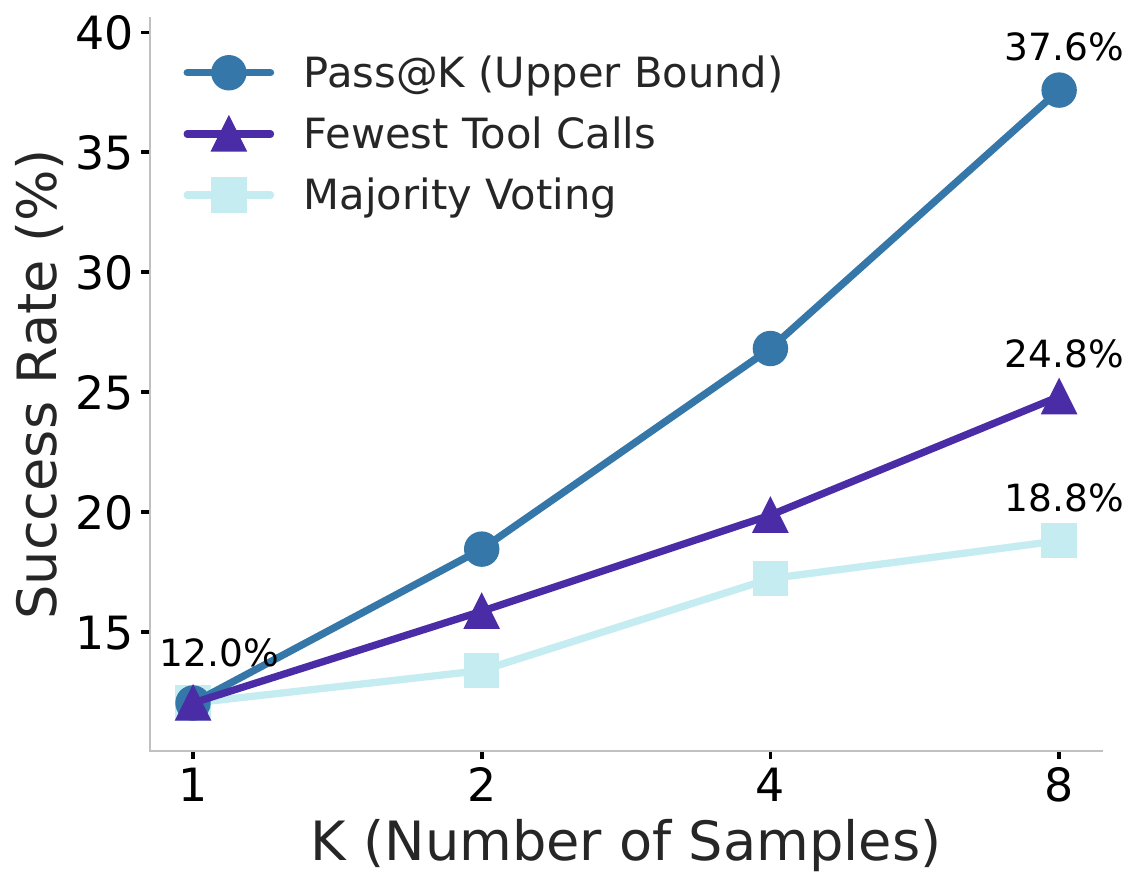}
            \vspace{-3mm}
            \label{fig:parallel-sampling}
        \end{minipage}
    }
    \caption{Test-time scaling results for \model-32B. \textit{Left} and \textit{Middle}: Performance vs. maximum tool calls on BrowseComp and BrowseComp-ZH (x-axis in log scale). \textit{Right}: Parallel sampling comparison on BrowseComp-266 (a randomly sampled subset), showing that selecting answers with the fewest tool calls outperforms majority voting. }
    \label{fig:inference_tool_scaling}
\end{figure*}

Figure \ref{fig:browsecomp_scaling} and \ref{fig:browsecomp_zh_scaling} show the impact of increasing the maximum number of tool calls on BrowseComp and BrowseComp-ZH. 
Performance improves steadily as the tool call budget grows. 
When the tool call limit reaches 16 or more, \model-32B trained with reinforcement learning clearly outperforms its SFT-only counterpart, demonstrating the benefit of RL for tool call scaling. 
The dotted line indicates the QwQ-32B baseline, which is relatively low on both datasets. 
Although QwQ-32B achieves about 15 points on BrowseComp-ZH without tool use, our model surpasses this baseline once the tool call budget exceeds 16. 

\paragraph{Parallel Sampling and Tool Call Voting. }

Beyond scaling the number of tool calls, we investigate how parallel sampling can further improve performance. 
As shown in Figure~\ref{fig:parallel-sampling}, majority voting, which selects the most frequent answer~\citep{wang2022self}, improves \model-32B performance on BrowseComp-266 from 12.0 to 18.8.
We further analyze the distribution of tool calls across parallel samples and observe that answers requiring fewer tool calls before submission tend to be more accurate.
This pattern likely occurs because the model stops earlier when confident in a good answer, whereas additional calls often reflect uncertainty and lead to less reliable results.
Based on this observation, we propose selecting the answer with the fewest tool calls, which achieves a substantial improvement from 12.0 to 24.8, approaching the theoretical upper bound of 37.6 (pass@8). 

\subsection{Ablation Study}
\label{sec:ablation}

\paragraph{Reward Ablation. }

\begin{figure*}[!htbp]
    \vspace{-5pt}
    \centering
    \subfloat[Format Reward Ablation]{
        \begin{minipage}{0.3\textwidth}
            \centering
            \includegraphics[width=\textwidth]{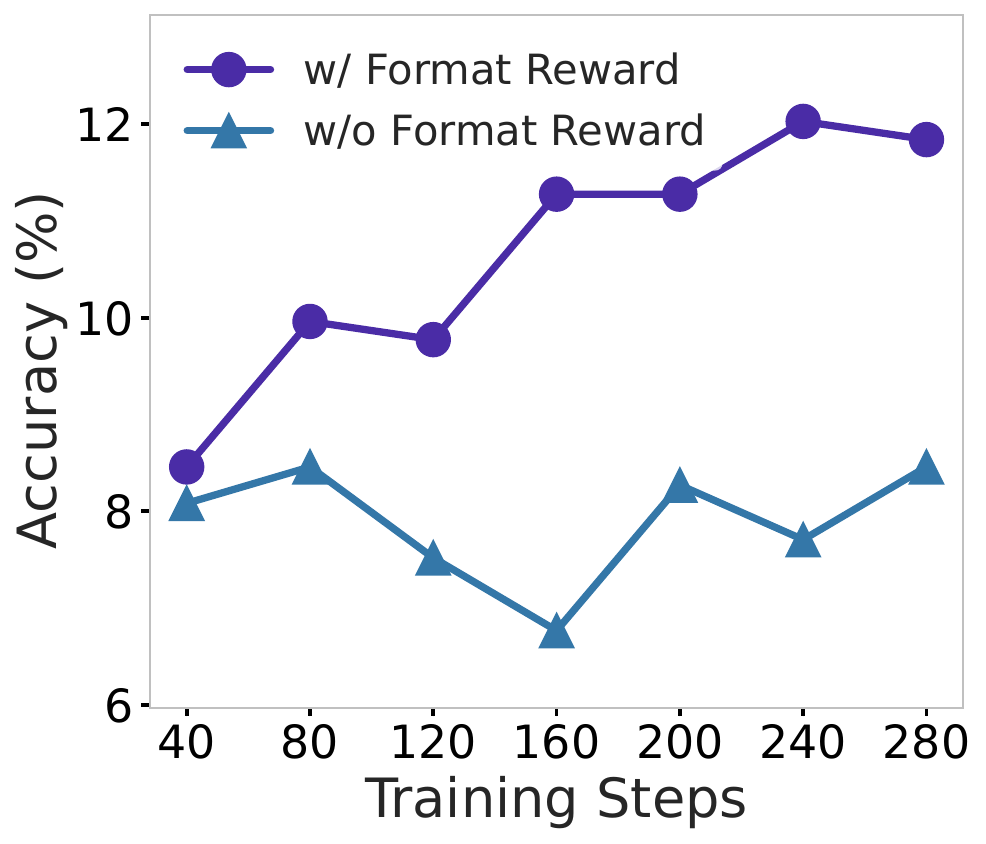}
            \label{fig:format_reward_ablation}
        \end{minipage}
    }\hfill
    \subfloat[Redundancy Penalty Ablation (Accuracy, $\uparrow$ higher is better)]{
        \begin{minipage}{0.3\textwidth}
            \centering
            \includegraphics[width=\textwidth]{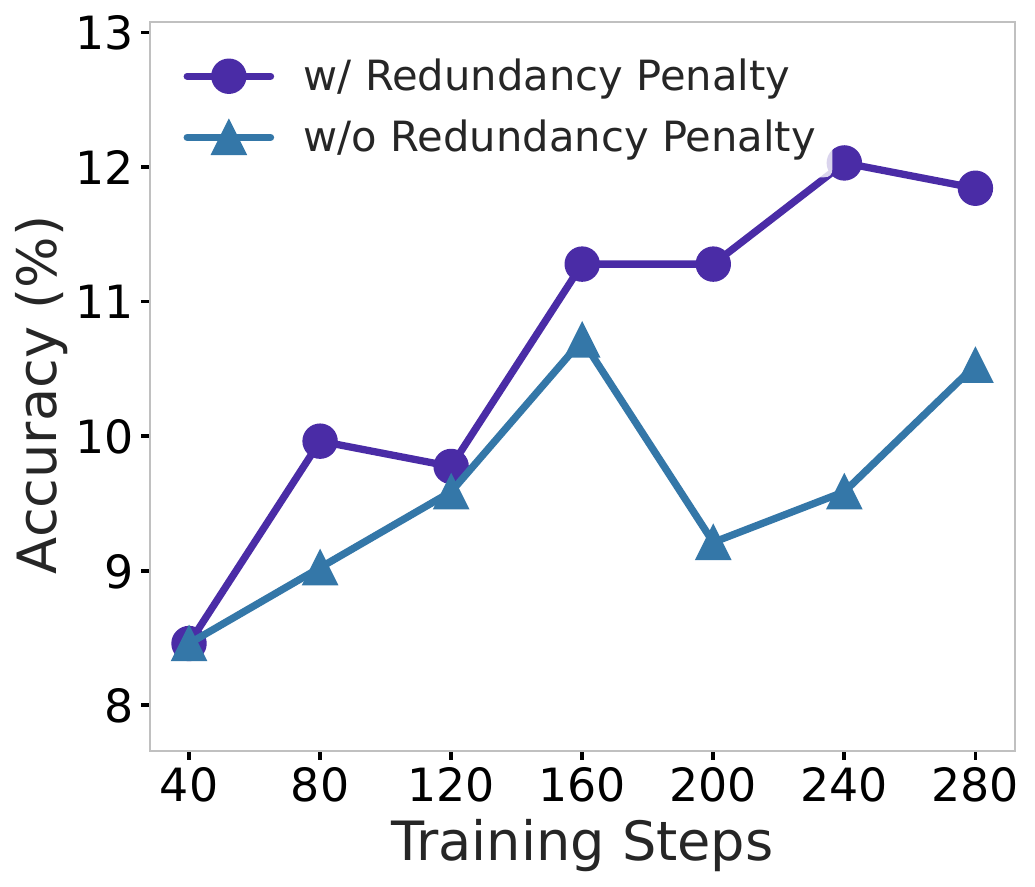}
            \label{fig:similarity_penalty_ablation_acc}
        \end{minipage}
    }\hfill
    \subfloat[Redundancy Penalty Ablation (Tool counts, $\downarrow$ lower is better)]{
        \begin{minipage}{0.3\textwidth}
            \centering
            \includegraphics[width=\textwidth]{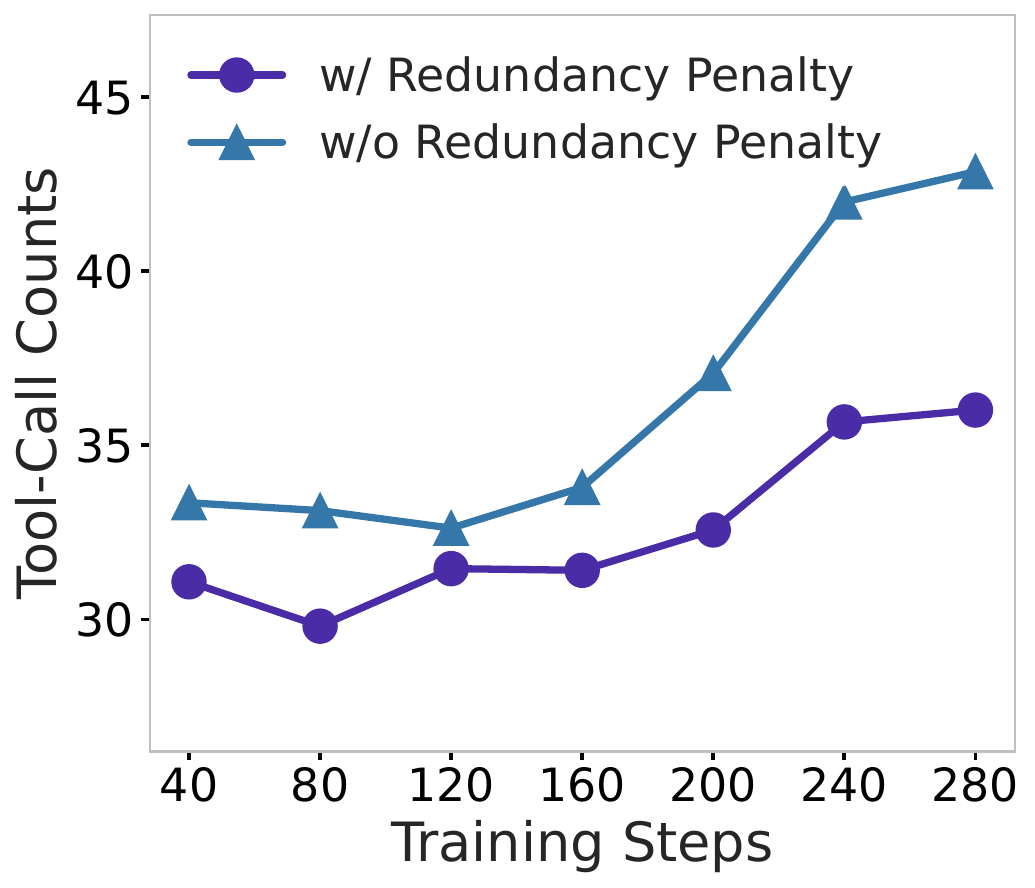}
            \label{fig:similarity_penalty_ablation_tool_call_count}
        \end{minipage}
    }
    \caption{Ablation of our reward design. All evaluations are on a sampled subset (BrowseComp-266). }
    \label{fig:reward_ablation}
    \vspace{-5pt}
\end{figure*}

We evaluate two components of our reward design under identical RL settings: the strict format reward and the redundancy penalty. 
All models are assessed on BrowseComp-266 (a randomly sampled subset) at regular intervals (every 40 steps from 40 to 240). 
In Figure~\ref{fig:format_reward_ablation}, removing the format reward yields a curve that stays near 8.0 with almost no improvement, while adding the format reward produces a steady upward trend that remains about 2 absolute points higher throughout training. 
In Figure~\ref{fig:similarity_penalty_ablation_acc} and Figure~\ref{fig:similarity_penalty_ablation_tool_call_count}, adding the redundancy penalty increases accuracy in the later training phase (about 20\%) and reduces tool call counts by roughly 14\% under the same conditions. 
Overall, the strict format reward accelerates and stabilizes learning, and the redundancy penalty prunes redundant searches, improving search efficiency without sacrificing performance. 

\paragraph{Synthetic Data Ablation. }

We ablate our synthetic deep search QA data across SFT and RL, using the same four benchmarks as the main experiments. 
We report accuracy (Acc) and average tool calls (\#Turn), where more tool calls indicate deeper search capability. As shown in Table~\ref{tab:ablation_combined}, the base QwQ-32B performs poorly with low accuracy and almost no tool use. 
SFT with HotpotQA trajectories gives only modest gains, while SFT with synthetic data brings clear improvements across all benchmarks, boosting both accuracy and tool usage. 
For RL, fixing the best SFT model, HotpotQA yields only minor gains without changing usage patterns, whereas synthetic data drives large gains on both metrics, especially on BrowseComp-266. 
In summary, our synthetic data proves essential for both training stages, boosting performance and enabling long-horizon deep search capabilities. 

\begin{table}[b]
  \centering
  \caption{Ablation study of different training data. For efficiency, we evaluate a subset of BrowseComp (BrowseComp-266), while the other three benchmarks are evaluated in full. }
  \label{tab:ablation_combined}
  \renewcommand{\arraystretch}{1.25}
  \begin{adjustbox}{max width=\textwidth}
    \begin{tabular}{l l *{4}{cc}}
      \toprule
      \multirow{2}{*}{\textbf{Backbone Model}} & \multirow{2}{*}{\textbf{Training Data}} & \multicolumn{2}{c}{\textbf{BrowseComp-266}} & \multicolumn{2}{c}{\textbf{BrowseComp-ZH}} & \multicolumn{2}{c}{\textbf{XBench-DeepSearch}} & \multicolumn{2}{c}{\textbf{SEAL-0}} \\
      \cmidrule(lr){3-4} \cmidrule(lr){5-6} \cmidrule(lr){7-8} \cmidrule(lr){9-10}
      & & \textbf{Acc} & \textbf{\#Turn} & \textbf{Acc} & \textbf{\#Turn} & \textbf{Acc} & \textbf{\#Turn} & \textbf{Acc} & \textbf{\#Turn} \\
      \midrule
      \multicolumn{10}{c}{\textit{Supervised Fine-tuning (SFT)}} \\
      \midrule
      \multirow{3}{*}{QwQ-32B} & – & 1.9 & 1.5 & 14.5 & 1.2 & 27.0 & 1.5 & 4.5 & 1.1 \\
      & + HotpotQA & 4.9 & 20.2 & 13.5 & 11.1 & 35.0 & 8.1 & 18.0 & 8.0 \\
      & + \textbf{our data} & \textbf{7.5} & \textbf{32.7} & \textbf{19.0} & \textbf{24.1} & \textbf{45.5} & \textbf{15.4} & \textbf{25.2} & \textbf{13.0} \\
      \midrule
      \multicolumn{10}{c}{\textit{Reinforcement Learning (RL) from the best SFT model}} \\
      \midrule
      \multirow{2}{*}{\model-32B (\textit{SFT only})} & + HotpotQA & 9.2 & 33.2 & 22.7 & 23.3 & 47.0 & 15.1 & 21.6 & 13.6 \\
      & + \textbf{our data} & \textbf{12.0} & \textbf{36.3} & \textbf{29.7} & \textbf{24.9} & \textbf{50.0} & \textbf{16.7} & \textbf{25.5} & \textbf{14.5} \\
      \bottomrule
    \end{tabular}
  \end{adjustbox}
\end{table}

\section{Additional Study: Semi-Automated i.i.d. Deep Search QA Synthesis for RL}
\label{sec:iid_webqa}

We perform an additional study to directly improve model performance on deep search benchmarks. 
Straightforwardly, we can construct i.i.d. QA pairs with BrowseComp, whose questions are so challenging that expert annotators have to spend hours solving them, ensuring that simple search strategies are ineffective. 
However, reaching the depth and breadth of BrowseComp requires heavy human effort in research, annotation, and data curation. 
To reduce annotation costs, we present a semi-automated framework. 
 
\paragraph{i.i.d. Data Synthesis.}
We adopt a semi-automated framework to reduce the burden on annotators, where each annotator is supported by the OpenAI o3 model \citep{o3} equipped with search capabilities and follows a four-stage process. 

First, based on the nine topical domains defined in BrowseComp, the annotator collaborates with the model to identify root domains that contain abundant factual and structured web content. 
Second, the annotator explores various linked pages within each root domain using the model’s navigation and search features, and selects verifiable entities along with their associated attributes. 
Third, the annotator conducts further targeted searches related to each selected entity and engages in multi-turn interactions with the model to construct new challenging multi-hop questions. These questions are carefully written to obscure key information while retaining verifiability. 
Fourth, the annotator uses the model to attempt to answer the synthetic question. If the answer is incorrect or if multiple plausible answers exist, the sample is discarded. 
The annotator also records the time taken by the model to arrive at an answer in order to identify questions that are more difficult and of higher quality. This workflow requires minimal prior knowledge from the annotator.

Through iterative model-guided discovery, question construction, and verification, annotators can efficiently produce complex, high-quality deep search QA pairs. The same procedure is applied to Chinese websites to enhance multilingual performance. 
As a result, we obtain a total of 2,997 English and 275 Chinese challenging deep search QA pairs. 

\paragraph{RL with i.i.d. Deep Search Data.}
We follow the same pipeline as Section~\ref{sec:experiment}, using SFT for cold-start and difficulty-based filtering to build a high-quality subset for RL, with all training configurations remaining identical to those before, except for the data.
Table~\ref{tab:iid-result} presents the performance after incorporating i.i.d. training data. 
Notably, the \model-32B-RL model achieves an accuracy of 20.8\% on the full BrowseComp benchmark, representing a 40\% improvement over the previous best score of 15.3\% and significantly outperforming open-source alternatives. 
Owing to the inclusion of Chinese content in the new training corpus, the new model also demonstrates considerable gains on Chinese-language benchmarks, namely BrowseComp-ZH and Xbench-DeepSearch.
Interestingly, performance on SEAL-0 remains largely unchanged, which we attribute to the dataset’s focus on recognizing and selecting among different search results, which is a challenge that highlights a key area for future model enhancement.

\begin{table}[htbp]
  \centering
  \caption{Effect of i.i.d. deep search QA data for \model. \model-32B Accuracy (\%) on 4 deep search benchmarks with and without i.i.d. data. \textbf{bold}: best performance; \underline{underline}: second best. }
  \label{tab:iid-result}
  \renewcommand{\arraystretch}{1.25}
  \begin{adjustbox}{max width=\textwidth}
    \begin{tabular}{lcccccc}
      \toprule
      \textbf{Model} & \textbf{data} &
      \textbf{BrowseComp} & \textbf{BrowseComp-ZH} & \textbf{Xbench-DeepSearch} & \textbf{SEAL-0} \\
      \midrule
      \model-32B (\textit{sft-only}) & KG data & 9.5 & 23.0 & 48.5 & 23.9 \\
      \model-32B & KG data & \underline{15.3} & \underline{29.7} & \underline{50.0} & \textbf{25.5} \\
      \midrule
      \blue{\model-32B (\textit{sft-only}) } & \blue{i.i.d data} & \blue{11.4} & \blue{26.6} & \blue{47.5} & \blue{22.5} \\
      \blue{\model}-32B & \blue{i.i.d data} & \blue{\textbf{22.2}} & \blue{\textbf{33.9}} & \blue{\textbf{56.0}} & \blue{\underline{23.0}} \\
      \bottomrule
    \end{tabular}
  \end{adjustbox}
\end{table}

\paragraph{Data Contamination Analysis.}

To ensure that the performance improvements are not the result of data leakage, we follow the contamination analysis protocol introduced in LLaMA 2~\citep{touvron2023llama2} and evaluate the Human-in-the-Loop dataset used for training. 
For each evaluation sample, we tokenize the input (excluding special tokens) and extract all contiguous 10-token n-grams. 
A token is considered contaminated if it appears in any n-gram also found in the training corpus. 
The contamination rate for a sample is defined as the proportion of contaminated tokens. Based on these rates, we categorize each sample into four non-exclusive subsets: Clean (less than 20\% contamination), Not Clean (20\% or more), Not Dirty (less than 80\%), and Dirty (80\% or more). 
As shown in Table~\ref{tab:contamination_rates}, more than 97\% of the samples in the dataset are classified as Clean, and there are no samples in the Dirty category. 
The results indicate that there is almost no test-data leakage in the constructed dataset for training. 

\begin{table}[ht]
  \centering
  \caption{Contamination analysis of BrowseComp evaluation samples using different synthetic data. 
  Each sample is categorized based on the proportion of overlapping n-grams with the training set.}
  \label{tab:contamination_rates}
  \renewcommand{\arraystretch}{1.25}
  \begin{tabular}{cccccc}
    \toprule
    \textbf{Data Type} &\textbf{Contamination Rate} & \textbf{Clean} & \textbf{Not Clean} & \textbf{Not Dirty} & \textbf{Dirty} \\
    \cmidrule(lr){1-6}
     KG & 2.6 & 99.0 & 1.0 & 100.0 & 0.0 \\
    i.i.d. & 3.4 & 97.7 & 2.3 & 100.0 & 0.0 \\
    \bottomrule
  \end{tabular}
\end{table}

Given the contamination analysis, both the KG data and i.i.d. data are adopted by the open GLM-4.5 models, which show strong performance on BrowseComp. 

\section{Related Work}

\paragraph{Reinforcement Learning for LLMs. }
Reinforcement learning has been central to advances in large language models. 
Early reinforcement learning from human feedback (RLHF) demonstrated how human preferences could align models with user intent~\citep{ouyang2022training}. 
Subsequent work shifted to verifiable reward signals to strengthen reasoning. 
Large-scale efforts such as OpenAI’s o1~\citep{o1} have empirically validated the effectiveness of verifiable-reward RL, while a wave of algorithmic improvements broadens the toolkit: 
GRPO~\citep{shao2024deepseekmath} removes the critic model to simplify and stabilize training;
DeepSeek’s R1~\citep{deepseek2025r1} builds on GRPO to achieve strong reasoning performance; 
and DAPO~\citep{yu2025dapo} introduces fine-grained RL adjustments for scalable, robust pipelines. 

\paragraph{Deep Search Agents.}
\citet{yao2023react} first introduced ReAct, a framework that interleaves explicit reasoning with actions to tackle complex tasks. 
Recent deep research agents, like DeepResearch~\citep{openai2025deepresearch} and Gemini Deep Research~\citep{gemini_deep_research}, have reached near-expert levels in information seeking and reasoning. 
Proprietary systems like DeepResearch~\citep{openai2025deepresearch} and Gemini Deep Research~\citep{gemini_deep_research} reach near-expert levels. 
Open-source efforts include reinforcement learning approaches (ReSearch~\citep{chen2025research}, Search-o1~\citep{li2025search}, WebThinker~\citep{li2025webthinker}, DeepResearcher~\citep{zheng2025deepresearcher}, Search-R1~\citep{jin2025search}, WebExplorer~\citep{liu2025webexplorer}, and WebShaper~\citep{tao2025webshaper}) that optimize tool use and retrieval, and framework-based systems (OpenDeepResearch~\citep{huggingface2025deepresearch}, TTD-DR~\citep{han2025deep}) that target long-form generation. 
A significant gap remains between open-source and proprietary models. 
\section{Conclusion}
\label{sec:conclusion}

We present \model that aligns deep reasoning with multi-turn web search through automated deep search QA synthesis and end-to-end multi-turn reinforcement learning. 
Our data pipeline generates ambiguity-rich, multi-hop questions with hidden cues, and our training introduces a redundancy penalty to encourage diverse and efficient search.
After the RL stage, \model-32B achieves 15.3\% accuracy on BrowseComp, setting a new competitive standard for open-source models while surpassing larger agents and multiple strong proprietary baselines. 
Analyses show that complex supervision and multi-turn RL jointly ground tool use, that performance scales with tool-call budgets and parallel sampling, and that skills learned on hard problems transfer to simpler settings. 
\section{Limitation}
Although the two challenging deep research QA data synthesis methods we proposed enable us to generate high-quality data with guaranteed difficulty and accuracy—helping \model-32B achieve competitive results among open-source models—the upper limit of difficulty remains significantly lower than datasets like BrowseComp. This limitation indirectly contributes to \model-32B substantially lower performance on BrowseComp compared to advanced models such as o3 with browsing. 
Additionally, our training approach, which primarily targets difficult data, has led to an "over-search" phenomenon in \model-32B. Determining optimal training steps and designing more appropriate reward mechanisms for the reinforcement learning stage represents an important area for future exploration. 

\bibliographystyle{plainnat}
\bibliography{references}

\begin{thebibliography}{57}
\providecommand{\natexlab}[1]{#1}
\providecommand{\url}[1]{\texttt{#1}}
\expandafter\ifx\csname urlstyle\endcsname\relax
  \providecommand{\doi}[1]{doi: #1}\else
  \providecommand{\doi}{doi: \begingroup \urlstyle{rm}\Url}\fi

\bibitem[Anthropic(2025{\natexlab{a}})]{anthropic2025extended-thinking}
Anthropic.
\newblock {Building with Extended Thinking in Claude 4 Models}.
\newblock \url{https://docs.anthropic.com/en/docs/build-with-claude/extended-thinking}, 2025{\natexlab{a}}.
\newblock Accessed July 16, 2025.

\bibitem[Anthropic(2025{\natexlab{b}})]{claude37}
Anthropic.
\newblock Claude 3.7 {Sonnet}, 2025{\natexlab{b}}.
\newblock URL \url{https://www.anthropic.com/news/claude-3-7-sonnet}.

\bibitem[Chen et~al.(2025)Chen, Li, Sun, Zhou, Zhu, Wang, Pan, Zhang, Chen, Yang, Zhou, and Chen]{chen2025research}
Mingyang Chen, Tianpeng Li, Haoze Sun, Yijie Zhou, Chenzheng Zhu, Haofen Wang, Jeff~Z. Pan, Wen Zhang, Huajun Chen, Fan Yang, Zenan Zhou, and Weipeng Chen.
\newblock {ReSearch}: Learning to reason with search for {LLMs} via reinforcement learning.
\newblock \emph{arXiv preprint arXiv:2503.19470}, 2025.

\bibitem[DeepSeek-AI et~al.(2025)DeepSeek-AI, Guo, Yang, Zhang, Song, and et~al.]{deepseek2025r1}
DeepSeek-AI, Daya Guo, Dejian Yang, Haowei Zhang, Junxiao Song, and et~al.
\newblock {DeepSeek-R1}: Incentivizing reasoning capability in {LLMs} via reinforcement learning.
\newblock \emph{arXiv preprint arXiv:2501.12948}, 2025.

\bibitem[Doubao(2025)]{doubao}
ByteDance Doubao.
\newblock Doubao, 2025.
\newblock URL \url{http://www.doubao.com/}.

\bibitem[Dubey et~al.(2024)Dubey, Jauhri, Pandey, Kadian, Al{-}Dahle, Letman, Mathur, Schelten, Yang, Fan, Goyal, Hartshorn, Yang, Mitra, Sravankumar, Korenev, Hinsvark, Rao, Zhang, Rodriguez, Gregerson, Spataru, Rozi{\`{e}}re, Biron, Tang, Chern, Caucheteux, Nayak, Bi, Marra, McConnell, Keller, Touret, Wu, Wong, Ferrer, Nikolaidis, Allonsius, Song, Pintz, Livshits, Esiobu, Choudhary, Mahajan, Garcia{-}Olano, Perino, Hupkes, Lakomkin, AlBadawy, Lobanova, Dinan, Smith, Radenovic, Zhang, Synnaeve, Lee, Anderson, Nail, Mialon, Pang, Cucurell, Nguyen, Korevaar, Xu, Touvron, Zarov, Ibarra, Kloumann, Misra, Evtimov, Copet, Lee, Geffert, Vranes, Park, Mahadeokar, Shah, van~der Linde, Billock, Hong, Lee, Fu, Chi, Huang, Liu, Wang, Yu, Bitton, Spisak, Park, Rocca, Johnstun, Saxe, Jia, Alwala, Upasani, Plawiak, Li, Heafield, Stone, and et~al.]{llama3}
Abhimanyu Dubey, Abhinav Jauhri, Abhinav Pandey, Abhishek Kadian, Ahmad Al{-}Dahle, Aiesha Letman, Akhil Mathur, Alan Schelten, Amy Yang, Angela Fan, Anirudh Goyal, Anthony Hartshorn, Aobo Yang, Archi Mitra, Archie Sravankumar, Artem Korenev, Arthur Hinsvark, Arun Rao, Aston Zhang, Aur{\'{e}}lien Rodriguez, Austen Gregerson, Ava Spataru, Baptiste Rozi{\`{e}}re, Bethany Biron, Binh Tang, Bobbie Chern, Charlotte Caucheteux, Chaya Nayak, Chloe Bi, Chris Marra, Chris McConnell, Christian Keller, Christophe Touret, Chunyang Wu, Corinne Wong, Cristian~Canton Ferrer, Cyrus Nikolaidis, Damien Allonsius, Daniel Song, Danielle Pintz, Danny Livshits, David Esiobu, Dhruv Choudhary, Dhruv Mahajan, Diego Garcia{-}Olano, Diego Perino, Dieuwke Hupkes, Egor Lakomkin, Ehab AlBadawy, Elina Lobanova, Emily Dinan, Eric~Michael Smith, Filip Radenovic, Frank Zhang, Gabriel Synnaeve, Gabrielle Lee, Georgia~Lewis Anderson, Graeme Nail, Gr{\'{e}}goire Mialon, Guan Pang, Guillem Cucurell, Hailey Nguyen, Hannah Korevaar, Hu~Xu, Hugo
  Touvron, Iliyan Zarov, Imanol~Arrieta Ibarra, Isabel~M. Kloumann, Ishan Misra, Ivan Evtimov, Jade Copet, Jaewon Lee, Jan Geffert, Jana Vranes, Jason Park, Jay Mahadeokar, Jeet Shah, Jelmer van~der Linde, Jennifer Billock, Jenny Hong, Jenya Lee, Jeremy Fu, Jianfeng Chi, Jianyu Huang, Jiawen Liu, Jie Wang, Jiecao Yu, Joanna Bitton, Joe Spisak, Jongsoo Park, Joseph Rocca, Joshua Johnstun, Joshua Saxe, Junteng Jia, Kalyan~Vasuden Alwala, Kartikeya Upasani, Kate Plawiak, Ke~Li, Kenneth Heafield, Kevin Stone, and et~al.
\newblock The {Llama} 3 herd of models.
\newblock \emph{CoRR}, abs/2407.21783, 2024.

\bibitem[Gao et~al.(2025)Gao, Fu, Xie, Xu, He, Mei, Zhu, and Wu]{gao2025beyond}
Jiaxuan Gao, Wei Fu, Minyang Xie, Shusheng Xu, Chuyi He, Zhiyu Mei, Banghua Zhu, and Yi~Wu.
\newblock Beyond ten turns: Unlocking long-horizon agentic search with large-scale asynchronous rl.
\newblock \emph{arXiv preprint arXiv:2508.07976}, 2025.

\bibitem[Gemini(2025)]{gemini_deep_research}
Gemini.
\newblock Gemini deep research.
\newblock \url{https://gemini.google/overview/deep-research}, 2025.

\bibitem[GLM et~al.(2024)GLM, Zeng, Xu, Wang, Zhang, Yin, Rojas, Feng, Zhao, Lai, et~al.]{glm2024chatglm}
Team GLM, Aohan Zeng, Bin Xu, Bowen Wang, Chenhui Zhang, Da~Yin, Diego Rojas, Guanyu Feng, Hanlin Zhao, Hanyu Lai, et~al.
\newblock Chatglm: A family of large language models from glm-130b to glm-4 all tools.
\newblock \emph{arXiv preprint arXiv:2406.12793}, 2024.

\bibitem[GLM-4.5 et~al.(2025)GLM-4.5, Zeng, Lv, Zheng, Hou, Chen, Xie, Wang, Yin, Zeng, Zhang, et~al.]{team2025glm45}
Team GLM-4.5, Aohan Zeng, Xin Lv, Qinkai Zheng, Zhenyu Hou, Bin Chen, Chengxing Xie, Cunxiang Wang, Da~Yin, Hao Zeng, Jiajie Zhang, et~al.
\newblock Glm-4.5: Agentic, reasoning, and coding (arc) foundation models.
\newblock \emph{arXiv preprint arXiv:2508.06471}, 2025.

\bibitem[Grok(2025)]{grok4}
Grok.
\newblock Grok 4.
\newblock \url{https://x.ai/news/grok-4}, 2025.

\bibitem[Guo et~al.(2025)Guo, Yang, Zhang, Song, Zhang, Xu, Zhu, Ma, Wang, Bi, et~al.]{guo2025deepseek}
Daya Guo, Dejian Yang, Haowei Zhang, Junxiao Song, Ruoyu Zhang, Runxin Xu, Qihao Zhu, Shirong Ma, Peiyi Wang, Xiao Bi, et~al.
\newblock Deepseek-r1: Incentivizing reasoning capability in llms via reinforcement learning.
\newblock \emph{arXiv preprint arXiv:2501.12948}, 2025.

\bibitem[Han et~al.(2025)Han, Chen, CuiZhu, Miculicich, Sun, Bi, Wen, Wan, Wen, Ma{\^\i}tre, et~al.]{han2025deep}
Rujun Han, Yanfei Chen, Zoey CuiZhu, Lesly Miculicich, Guan Sun, Yuanjun Bi, Weiming Wen, Hui Wan, Chunfeng Wen, Sol{\`e}ne Ma{\^\i}tre, et~al.
\newblock Deep researcher with test-time diffusion.
\newblock \emph{arXiv preprint arXiv:2507.16075}, 2025.

\bibitem[Ho et~al.(2020)Ho, Nguyen, Sugawara, and Aizawa]{ho2020constructing}
Xanh Ho, Anh-Khoa~Duong Nguyen, Saku Sugawara, and Akiko Aizawa.
\newblock Constructing a multi-hop qa dataset for comprehensive evaluation of reasoning steps.
\newblock \emph{arXiv preprint arXiv:2011.01060}, 2020.

\bibitem[Hou et~al.(2025)Hou, Lv, Lu, Zhang, Li, Yao, Li, Tang, and Dong]{hou2025advancing}
Zhenyu Hou, Xin Lv, Rui Lu, Jiajie Zhang, Yujiang Li, Zijun Yao, Juanzi Li, Jie Tang, and Yuxiao Dong.
\newblock Advancing language model reasoning through reinforcement learning and inference scaling.
\newblock \emph{arXiv preprint arXiv:2501.11651}, 2025.

\bibitem[{Hugging Face}(2025)]{huggingface2025deepresearch}
{Hugging Face}.
\newblock Open-source deep research -- freeing our search agents.
\newblock Hugging Face Blog, 2025.
\newblock URL: \url{https://huggingface.co/blog/open-deep-research}.

\bibitem[Ji et~al.(2021)Ji, Pan, Cambria, Marttinen, and Yu]{ji2021survey}
Shaoxiong Ji, Shirui Pan, Erik Cambria, Pekka Marttinen, and Philip~S Yu.
\newblock A survey on knowledge graphs: Representation, acquisition, and applications.
\newblock \emph{IEEE transactions on neural networks and learning systems}, 33\penalty0 (2):\penalty0 494--514, 2021.

\bibitem[Jin et~al.(2025)Jin, Zeng, Yue, Yoon, Arik, Wang, Zamani, and Han]{jin2025search}
Bowen Jin, Hansi Zeng, Zhenrui Yue, Jinsung Yoon, Sercan Arik, Dong Wang, Hamed Zamani, and Jiawei Han.
\newblock Search-r1: Training llms to reason and leverage search engines with reinforcement learning.
\newblock \emph{arXiv preprint arXiv:2503.09516}, 2025.

\bibitem[Jina.ai(2025)]{jina}
Jina.ai.
\newblock Jina, 2025.
\newblock URL \url{https://jina.ai/}.

\bibitem[Krishna et~al.(2024)Krishna, Krishna, Mohananey, Schwarcz, Stambler, Upadhyay, and Faruqui]{krishna2024fact}
Satyapriya Krishna, Kalpesh Krishna, Anhad Mohananey, Steven Schwarcz, Adam Stambler, Shyam Upadhyay, and Manaal Faruqui.
\newblock Fact, fetch, and reason: A unified evaluation of retrieval-augmented generation.
\newblock \emph{arXiv preprint arXiv:2409.12941}, 2024.

\bibitem[Li et~al.(2025{\natexlab{a}})Li, Zhang, Yin, Zhang, Ou, Wu, Yin, Li, Tao, Wang, Shen, Zhang, Zhang, Wu, Jiang, Yan, Xie, Huang, and Zhou]{li2025websailornavigatingsuperhumanreasoning}
Kuan Li, Zhongwang Zhang, Huifeng Yin, Liwen Zhang, Litu Ou, Jialong Wu, Wenbiao Yin, Baixuan Li, Zhengwei Tao, Xinyu Wang, Weizhou Shen, Junkai Zhang, Dingchu Zhang, Xixi Wu, Yong Jiang, Ming Yan, Pengjun Xie, Fei Huang, and Jingren Zhou.
\newblock Websailor: Navigating super-human reasoning for web agent, 2025{\natexlab{a}}.
\newblock URL \url{https://arxiv.org/abs/2507.02592}.

\bibitem[Li et~al.(2025{\natexlab{b}})Li, Dong, Jin, Zhang, Zhou, Zhu, Zhang, and Dou]{li2025search}
Xiaoxi Li, Guanting Dong, Jiajie Jin, Yuyao Zhang, Yujia Zhou, Yutao Zhu, Peitian Zhang, and Zhicheng Dou.
\newblock Search-o1: Agentic search-enhanced large reasoning models.
\newblock \emph{arXiv preprint arXiv:2501.05366}, 2025{\natexlab{b}}.

\bibitem[Li et~al.(2025{\natexlab{c}})Li, Jin, Dong, Qian, Zhu, Wu, Wen, and Dou]{li2025webthinker}
Xiaoxi Li, Jiajie Jin, Guanting Dong, Hongjin Qian, Yutao Zhu, Yongkang Wu, Ji-Rong Wen, and Zhicheng Dou.
\newblock {WebThinker}: Empowering large reasoning models with deep research capability.
\newblock \emph{arXiv preprint arXiv:2504.21776}, 2025{\natexlab{c}}.

\bibitem[Liu et~al.(2025)Liu, Li, Zhang, Li, Chen, Ji, Cheng, Wu, Du, Xu, et~al.]{liu2025webexplorer}
Junteng Liu, Yunji Li, Chi Zhang, Jingyang Li, Aili Chen, Ke~Ji, Weiyu Cheng, Zijia Wu, Chengyu Du, Qidi Xu, et~al.
\newblock Webexplorer: Explore and evolve for training long-horizon web agents.
\newblock \emph{arXiv preprint arXiv:2509.06501}, 2025.

\bibitem[{OpenAI}(2024{\natexlab{a}})]{gpt4o}
{OpenAI}.
\newblock Hello {GPT-4o}, 2024{\natexlab{a}}.
\newblock URL \url{https://openai.com/index/hello-gpt-4o/}.

\bibitem[{OpenAI}(2024{\natexlab{b}})]{o1}
{OpenAI}.
\newblock Learning to reason with {LLMs}, 2024{\natexlab{b}}.
\newblock URL \url{https://openai.com/index/learning-to-reason-with-llms/}.

\bibitem[OpenAI(2025)]{o3}
OpenAI.
\newblock Introducing openai o3 and o4-mini, 2025.
\newblock URL \url{https://openai.com/index/introducing-o3-and-o4-mini/}.

\bibitem[{OpenAI}(2025)]{openai2025deepresearch}
{OpenAI}.
\newblock Deep research: Autonomous web-research agent.
\newblock \url{https://openai.com/index/introducing-deep-research/}, 2025.

\bibitem[Ouyang et~al.(2022)Ouyang, Wu, Jiang, Almeida, Wainwright, Mishkin, Zhang, Agarwal, Slama, Ray, et~al.]{ouyang2022training}
Long Ouyang, Jeff Wu, Xu~Jiang, Diogo Almeida, Carroll~L Wainwright, Pamela Mishkin, Chong Zhang, Sandhini Agarwal, Katarina Slama, Alex Ray, et~al.
\newblock Training language models to follow instructions with human feedback.
\newblock In \emph{Proceedings of the 36th International Conference on Neural Information Processing Systems}, pages 27730--27744, 2022.

\bibitem[Petroni et~al.(2020)Petroni, Piktus, Fan, Lewis, Yazdani, De~Cao, Thorne, Jernite, Karpukhin, Maillard, et~al.]{petroni2020kilt}
Fabio Petroni, Aleksandra Piktus, Angela Fan, Patrick Lewis, Majid Yazdani, Nicola De~Cao, James Thorne, Yacine Jernite, Vladimir Karpukhin, Jean Maillard, et~al.
\newblock Kilt: a benchmark for knowledge intensive language tasks.
\newblock \emph{arXiv preprint arXiv:2009.02252}, 2020.

\bibitem[Pham et~al.(2025)Pham, Nguyen, Zunjare, Chen, Tseng, and Vu]{pham2025sealqaraisingbarreasoning}
Thinh Pham, Nguyen Nguyen, Pratibha Zunjare, Weiyuan Chen, Yu-Min Tseng, and Tu~Vu.
\newblock Sealqa: Raising the bar for reasoning in search-augmented language models, 2025.
\newblock URL \url{https://arxiv.org/abs/2506.01062}.

\bibitem[Press et~al.(2022)Press, Zhang, Min, Schmidt, Smith, and Lewis]{press2022measuring}
Ofir Press, Muru Zhang, Sewon Min, Ludwig Schmidt, Noah~A Smith, and Mike Lewis.
\newblock Measuring and narrowing the compositionality gap in language models.
\newblock \emph{arXiv preprint arXiv:2210.03350}, 2022.

\bibitem[Real and Vargas(1996)]{real1996probabilistic}
Raimundo Real and Juan~M Vargas.
\newblock The probabilistic basis of jaccard's index of similarity.
\newblock \emph{Systematic Biology}, 45\penalty0 (3):\penalty0 380--385, 1996.
\newblock \doi{10.1093/sysbio/45.3.380}.

\bibitem[Serper()]{serper2025}
Serper.
\newblock Serper: Google search api.
\newblock \url{https://serper.dev}, 2025.

\bibitem[Shao et~al.(2024)Shao, Wang, Zhu, Xu, Song, Zhang, Li, Wu, and Guo]{shao2024deepseekmath}
Zhihong Shao, Peiyi Wang, Qihao Zhu, Runxin Xu, Junxiao Song, Mingchuan Zhang, YK~Li, Yu~Wu, and Daya Guo.
\newblock Deepseekmath: Pushing the limits of mathematical reasoning in open language models.
\newblock \emph{arXiv preprint arXiv:2402.03300}, 2024.

\bibitem[Song et~al.(2025)Song, Jiang, Min, Chen, Chen, Zhao, Fang, and Wen]{song2025r1}
Huatong Song, Jinhao Jiang, Yingqian Min, Jie Chen, Zhipeng Chen, Wayne~Xin Zhao, Lei Fang, and Ji-Rong Wen.
\newblock R1-searcher: Incentivizing the search capability in llms via reinforcement learning.
\newblock \emph{arXiv preprint arXiv:2503.05592}, 2025.

\bibitem[Tang et~al.(2012)Tang, Zhang, Yao, Yu, Li, Zhang, Su, Wang, and Yang]{tang2012arnetminer}
Jie Tang, Jing Zhang, Limin Yao, Zhong Yu, Juanzi Li, Li~Zhang, Zhong Su, Dong Wang, and Qiang Yang.
\newblock Arnetminer: A comprehensive academic search and mining platform.
\newblock In \emph{Proceedings of the 18th ACM SIGKDD international conference on Knowledge discovery and data mining}, pages 1231--1239. ACM, 2012.

\bibitem[Tao et~al.(2025)Tao, Wu, Yin, Zhang, Li, Shen, Li, Zhang, Wang, Jiang, et~al.]{tao2025webshaper}
Zhengwei Tao, Jialong Wu, Wenbiao Yin, Junkai Zhang, Baixuan Li, Haiyang Shen, Kuan Li, Liwen Zhang, Xinyu Wang, Yong Jiang, et~al.
\newblock Webshaper: Agentically data synthesizing via information-seeking formalization.
\newblock \emph{arXiv preprint arXiv:2507.15061}, 2025.

\bibitem[Team et~al.(2023)Team, Anil, Borgeaud, Alayrac, Yu, Soricut, Schalkwyk, Dai, Hauth, Millican, et~al.]{team2023gemini}
Gemini Team, Rohan Anil, Sebastian Borgeaud, Jean-Baptiste Alayrac, Jiahui Yu, Radu Soricut, Johan Schalkwyk, Andrew~M Dai, Anja Hauth, Katie Millican, et~al.
\newblock Gemini: a family of highly capable multimodal models.
\newblock \emph{arXiv preprint arXiv:2312.11805}, 2023.

\bibitem[Team(2025)]{qwq32b}
Qwen Team.
\newblock Qwq-32b: Embracing the power of reinforcement learning, March 2025.
\newblock URL \url{https://qwenlm.github.io/blog/qwq-32b/}.

\bibitem[Touvron et~al.(2023)Touvron, Martin, Stone, Albert, Almahairi, Babaei, Bashlykov, Batra, Bhargava, Bhosale, et~al.]{touvron2023llama2}
Hugo Touvron, Louis Martin, Kevin Stone, Peter Albert, Amjad Almahairi, Yasmine Babaei, Nikolay Bashlykov, Soumya Batra, Prajjwal Bhargava, Shruti Bhosale, et~al.
\newblock Llama 2: Open foundation and fine-tuned chat models.
\newblock \emph{arXiv preprint arXiv:2307.09288}, 2023.

\bibitem[Trivedi et~al.(2022)Trivedi, Balasubramanian, Khot, and Sabharwal]{trivedi2022musique}
Harsh Trivedi, Niranjan Balasubramanian, Tushar Khot, and Ashish Sabharwal.
\newblock Musique: Multihop questions via single-hop question composition.
\newblock \emph{Transactions of the Association for Computational Linguistics}, 10:\penalty0 539--554, 2022.

\bibitem[Vassoyan et~al.(2025)Vassoyan, Beau, and Plaud]{Vassoyan2025IgnoreKL}
Jean Vassoyan, Nathanaël Beau, and Roman Plaud.
\newblock Ignore the kl penalty! boosting exploration on critical tokens to enhance rl fine-tuning.
\newblock \emph{arXiv preprint arXiv:2502.06533}, 2025.

\bibitem[Wang et~al.(2022)Wang, Wei, Schuurmans, Le, Chi, Narang, Chowdhery, and Zhou]{wang2022self}
Xuezhi Wang, Jason Wei, Dale Schuurmans, Quoc Le, Ed~Chi, Sharan Narang, Aakanksha Chowdhery, and Denny Zhou.
\newblock Self-consistency improves chain of thought reasoning in language models.
\newblock \emph{arXiv preprint arXiv:2203.11171}, 2022.

\bibitem[Wei et~al.(2022)Wei, Wang, Schuurmans, Bosma, Ichter, Xia, Chi, Le, and Zhou]{wei2022chain}
Jason Wei, Xuezhi Wang, Dale Schuurmans, Maarten Bosma, Brian Ichter, Fei Xia, Ed~H Chi, Quoc~V Le, and Denny Zhou.
\newblock Chain-of-thought prompting elicits reasoning in large language models.
\newblock In \emph{Proceedings of the 36th International Conference on Neural Information Processing Systems}, pages 24824--24837, 2022.

\bibitem[Wei et~al.(2025)Wei, Sun, Papay, McKinney, Han, Fulford, Chung, Passos, Fedus, and Glaese]{wei2025browsecomp}
Jason Wei, Zhiqing Sun, Spencer Papay, Scott McKinney, Jeffrey Han, Isa Fulford, Hyung~Won Chung, Alex~Tachard Passos, William Fedus, and Amelia Glaese.
\newblock Browsecomp: A simple yet challenging benchmark for browsing agents.
\newblock \emph{arXiv preprint arXiv:2504.12516}, 2025.

\bibitem[Wu et~al.(2025{\natexlab{a}})Wu, Li, Fang, Yin, Zhang, Tao, Zhang, Xi, Jiang, Xie, et~al.]{wu2025webdancer}
Jialong Wu, Baixuan Li, Runnan Fang, Wenbiao Yin, Liwen Zhang, Zhengwei Tao, Dingchu Zhang, Zekun Xi, Yong Jiang, Pengjun Xie, et~al.
\newblock Webdancer: Towards autonomous information seeking agency.
\newblock \emph{arXiv preprint arXiv:2505.22648}, 2025{\natexlab{a}}.

\bibitem[Wu et~al.(2025{\natexlab{b}})Wu, Yin, Jiang, Wang, Xi, Fang, Zhang, He, Zhou, Xie, and Huang]{wu2025webwalkerbenchmarkingllmsweb}
Jialong Wu, Wenbiao Yin, Yong Jiang, Zhenglin Wang, Zekun Xi, Runnan Fang, Linhai Zhang, Yulan He, Deyu Zhou, Pengjun Xie, and Fei Huang.
\newblock Webwalker: Benchmarking llms in web traversal, 2025{\natexlab{b}}.
\newblock URL \url{https://arxiv.org/abs/2501.07572}.

\bibitem[x.ai(2025)]{grok}
x.ai.
\newblock Grok 3 beta — the age of reasoning agents, 2025.
\newblock URL \url{https://x.ai/news/grok-3}.

\bibitem[Xbench-Team(2025)]{xbench}
Xbench-Team.
\newblock Xbench-deepsearch, 2025.
\newblock URL \url{https://xbench.org/agi/aisearch}.

\bibitem[Yang et~al.(2018)Yang, Qi, Zhang, Bengio, Cohen, Salakhutdinov, and Manning]{yang2018hotpotqa}
Zhilin Yang, Peng Qi, Saizheng Zhang, Yoshua Bengio, William~W Cohen, Ruslan Salakhutdinov, and Christopher~D Manning.
\newblock Hotpotqa: A dataset for diverse, explainable multi-hop question answering.
\newblock \emph{arXiv preprint arXiv:1809.09600}, 2018.

\bibitem[Yao et~al.(2023)Yao, Zhao, Yu, Du, Shafran, Narasimhan, and Cao]{yao2023react}
Shunyu Yao, Jeffrey Zhao, Dian Yu, Nan Du, Izhak Shafran, Karthik Narasimhan, and Yuan Cao.
\newblock React: Synergizing reasoning and acting in language models.
\newblock In \emph{International Conference on Learning Representations (ICLR)}, 2023.

\bibitem[Yu et~al.(2025)Yu, Zhang, Zhu, Yuan, Zuo, Yue, Dai, Fan, Liu, Liu, et~al.]{yu2025dapo}
Qiying Yu, Zheng Zhang, Ruofei Zhu, Yufeng Yuan, Xiaochen Zuo, Yu~Yue, Weinan Dai, Tiantian Fan, Gaohong Liu, Lingjun Liu, et~al.
\newblock Dapo: An open-source llm reinforcement learning system at scale.
\newblock \emph{arXiv preprint arXiv:2503.14476}, 2025.

\bibitem[Zheng et~al.(2023)Zheng, Chiang, Sheng, Zhuang, Wu, Zhuang, Lin, Li, Li, Xing, et~al.]{zheng2023judging}
Lianmin Zheng, Wei-Lin Chiang, Ying Sheng, Siyuan Zhuang, Zhanghao Wu, Yonghao Zhuang, Zi~Lin, Zhuohan Li, Dacheng Li, Eric Xing, et~al.
\newblock Judging llm-as-a-judge with mt-bench and chatbot arena.
\newblock \emph{Advances in Neural Information Processing Systems}, 36:\penalty0 46595--46623, 2023.

\bibitem[Zheng et~al.(2025)Zheng, Fu, Hu, and et~al.]{zheng2025deepresearcher}
Yuxiang Zheng, Dayuan Fu, Xiangkun Hu, and et~al.
\newblock {DeepResearcher}: Scaling deep research via reinforcement learning in real-world environments.
\newblock \emph{arXiv preprint arXiv:2504.03160}, 2025.

\bibitem[Zhou et~al.(2025)Zhou, Leon, Ying, Zhang, Shao, Ye, Chong, Jin, Xie, Cao, et~al.]{zhou2025browsecomp}
Peilin Zhou, Bruce Leon, Xiang Ying, Can Zhang, Yifan Shao, Qichen Ye, Dading Chong, Zhiling Jin, Chenxuan Xie, Meng Cao, et~al.
\newblock Browsecomp-zh: Benchmarking web browsing ability of large language models in chinese.
\newblock \emph{arXiv preprint arXiv:2504.19314}, 2025.

\bibitem[Zhu et~al.(2025)Zhu, Xie, Lv, and slime Contributors]{slime_github}
Zilin Zhu, Chengxing Xie, Xin Lv, and slime Contributors.
\newblock slime: An llm post-training framework for rl scaling.
\newblock \url{https://github.com/THUDM/slime}, 2025.
\newblock GitHub repository. Corresponding author: Xin Lv.

\end{thebibliography}

\newpage
\appendix

\newpage
\appendix
\section{Generalization on Simple Search Tasks}
\label{sec:generalization}

\begin{figure}[!h]
  \centering
  \includegraphics[width=0.6\columnwidth]{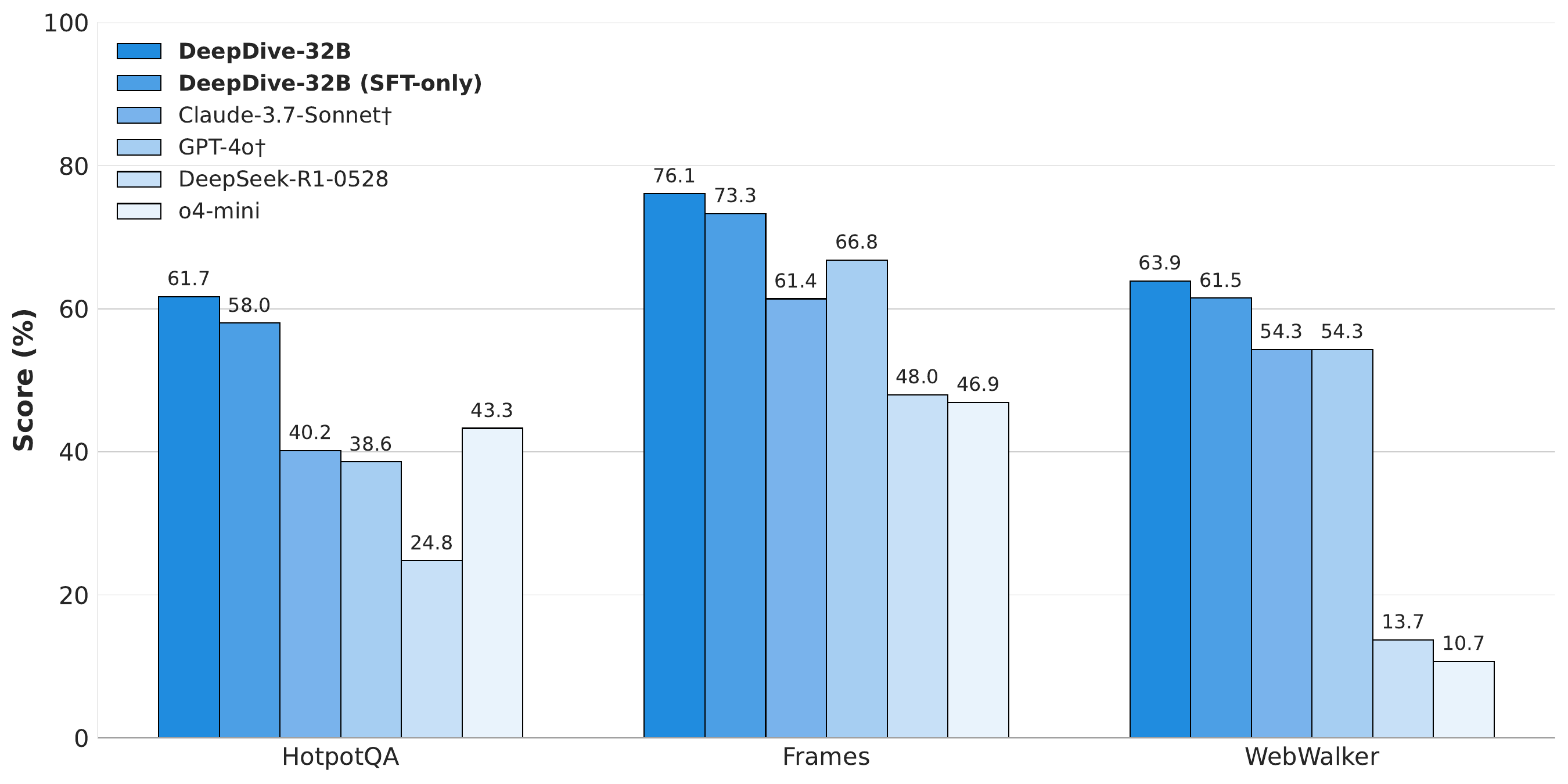}
  \caption{\model generalization on simple search benchmarks. }
  \label{fig:easy_task_generalization}
\end{figure}

While \model is trained on synthetic data based on knowledge graphs for challenging tasks like BrowseComp and BrowseComp-ZH, we evaluate its performance on simpler search benchmarks: HotpotQA \citep{yang2018hotpotqa}, Frames \citep{krishna2024fact}, and WebWalker \citep{wu2025webwalkerbenchmarkingllmsweb}, which involve more direct, less ambiguous questions. 
We compare \model against two non-search models (o4-mini and DeepSeek-R1-0528) and two proprietary search-enabled models using the same search engine. 
We evaluate 512 randomly selected HotpotQA questions and full test sets for other benchmarks. 
Figure~\ref{fig:easy_task_generalization} shows that both \model-32B (SFT-only) and \model-32B outperform all baselines, with reinforcement learning providing additional improvements across all benchmarks. 
These results confirm \model's strong generalization and search capabilities. 

\section{Case Study}
\paragraph{Reinforcement Learning Reshapes the Model’s Search Strategy}

\begin{figure}[!htbp]
  \centering
  \includegraphics[width=0.8\textwidth]{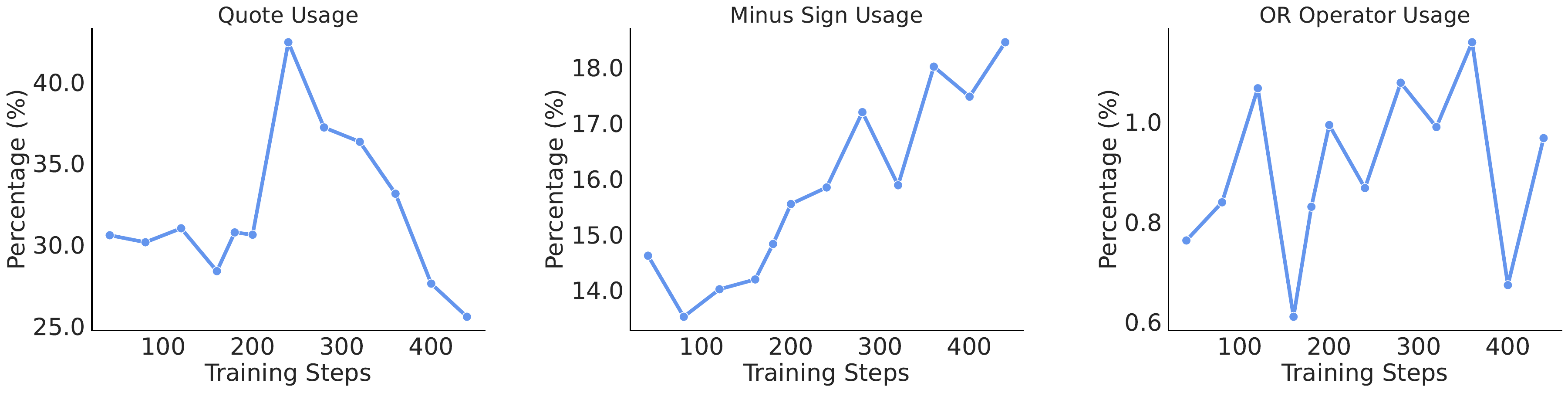}
  \caption{Evolution of \textit{Quote Usage}, \textit{Minus Usage} and \textit{OR Usage} over RL training steps on BrowseComp-266.}
  \label{fig:advance_search_ratio}
\end{figure}

Based on the sustained performance improvement on the BrowseComp-266 evaluation set during RL training, we study the model’s search behavior because most of its actions involve issuing retrieval queries. Similar to human interaction with contemporary search engines like Google Search, which allow exact match quoting, logical OR aggregation and term exclusion with a leading minus sign, the retrieval interface used during training and evaluation supports these same advanced features. We therefore collected every query generated by the model when solving the evaluation set and calculated three metrics: 
(1) \textit{Quote Usage}: the fraction of queries containing double quotes for exact phrase matching; 
(2) \textit{Minus Usage}: the fraction of queries containing a leading minus sign to exclude terms. 
(3) \textit{OR Usage}: the fraction of queries containing the OR operator to combine alternative terms; 
The evolution of these metrics over training steps is plotted in Figure~\ref{fig:advance_search_ratio}.

From Figure~\ref{fig:advance_search_ratio}, we observe that Quote Usage increases from around 30\% to 40\% at the early stage of training, then gradually decreases to below 25\%. 
OR Usage steadily grows from approximately 2\% to 8\%. 
In contrast, Minus Usage continues to rise from 14\% to 18\% throughout the training. This trend suggests that the model initially learns to adopt the quoting strategy early in reinforcement learning, but its advantage becomes less prominent over time, leading to a decline in usage. 
Meanwhile, the model steadily improves its ability to use minus operators, and OR Usage remains stable between 0.8\% and 1\%, indicating limited but consistent application.

\end{document}